\pdfoutput=1
\RequirePackage{fix-cm}

\documentclass[10pt,twocolumn,letterpaper]{article}
\usepackage[pagenumbers]{cvpr}
\usepackage{amsthm}
\usepackage{multirow}
\usepackage{colortbl}
\newcommand{\second}[1]{\underline{#1}}
\definecolor{oursrow}{gray}{0.92}
\definecolor{cvprblue}{rgb}{0.21,0.49,0.74}

\newcommand{\clr}{\textup{\textsc{CLR}}}
\newtheorem{proposition}{Proposition}

\usepackage{tabularx}


\usepackage[breaklinks,colorlinks,allcolors=cvprblue]{hyperref}
\usepackage[capitalize]{cleveref}
\hypersetup{
  pdftitle={Calibrated Reading of Latent Reasoning and Communication},
  pdfauthor={Zixi Wang, Gaohao Zhou, Yubo Huang, Xin Lai, Yushe Cao, Jingzehua Xu, Shuai Zhang}
}
\title{Calibrated Reading of Latent Reasoning and Communication}
\author{Zixi Wang\textsuperscript{1}\quad
Gaohao Zhou\textsuperscript{2}\quad
Yubo Huang\textsuperscript{3,*}\quad
Xin Lai\textsuperscript{3}\\
Yushe Cao\textsuperscript{4}\quad
Jingzehua Xu\textsuperscript{5}\quad
Shuai Zhang\textsuperscript{6}\\[4pt]
{\small\textsuperscript{1}University of Electronic Science and Technology of China}\\
{\small\textsuperscript{2}AI2Robotics\quad
\textsuperscript{3}Southwest Jiaotong University}\\
{\small\textsuperscript{4}Tsinghua University\quad
\textsuperscript{5}The University of Hong Kong}\\
{\small\textsuperscript{6}New Jersey Institute of Technology}\\[2pt]
{\small\textsuperscript{*}Corresponding author: \texttt{ybforever@my.swjtu.edu.cn}}
}
\date{}

\begin{document}
\maketitle
\begin{abstract}
Latent reasoning and key--value communication store intermediate information in continuous representations, limiting direct inspection of model behavior. Existing activation readers trade the low cost of task-specific probes against the flexibility of language-model decoders. We propose the Calibrated Latent Reader, \clr{}, a compact transformer that predicts reasoning outcomes and message content through task-conditioned interactions between states and candidate answers. Source adapters support transfer across models, while conformal calibration provides coverage and false-alarm guarantees under exchangeability. On two CODI models, \clr{} improves accuracy over MLP and attention probes by 4--18 percentage points for final-answer prediction and reference-value membership. On CODI-GPT-2, it achieves accuracy comparable to a 1.7B LatentQA-based verbalizer using 437--648 times fewer inference FLOPs. On synthetic Qwen3-4B communications, comparing decoded facts with public commitments detects 93.1\% of message substitutions at a nominal 5\% false-positive rate.
\end{abstract}

\section{Introduction}
\label{sec:intro}
\begin{figure}[t]
\centering
\includegraphics[width=\linewidth]{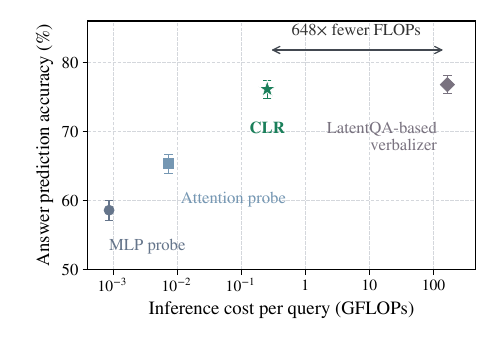}
\caption{Prediction accuracy versus inference cost on CODI-GPT-2. The task predicts the source's final answer among four candidates. Points show three-seed test accuracy; bars give 95\% intervals. FLOPs are measured per query at batch size one. \clr{} approaches the verbalizer's accuracy with $648\times$ fewer FLOPs.}
\label{fig:teaser}
\end{figure}

Chain-of-thought reasoning expands computation through intermediate text~\cite{wei2022chain}. Continuous reasoning instead allocates computation in hidden states. Coconut recurrently feeds hidden states into the model, CODI distills textual reasoning into continuous thoughts, and Huginn increases computation through recurrent depth~\cite{hao2024training,shen2025codi,geiping2026scaling}. A similar shift occurs in multi-agent systems that transmit hidden states or key--value caches instead of text~\cite{zou2025latent,fu2026cache}. These methods make intermediate computation less accessible to the textual monitors studied for reasoning models~\cite{baker2025monitoring,korbak2025chain}. Inspecting their behavior therefore requires reading the information carried by latent states.

Activation reading is commonly approached through probes or language-model decoders. Probes learn a mapping from a representation to a predefined property~\cite{alain2016understanding,belinkov2022probing}. Patchscopes and SelfIE instead use a language model to interpret activations, while LatentQA trains a decoder to answer questions about them~\cite{ghandeharioun2024patchscopes,chen2024selfie,pan2026latentqa}. Generative readers provide a flexible interface, but invoking another large model for each inspection adds substantial computation. Their outputs may also draw on the decoder's knowledge, requiring controls that distinguish information supplied by the target states from information available to the reader independently~\cite{li2025activation}.

Many monitoring decisions depend on a small set of alternatives. Before a reasoning model finishes, a monitor may need to identify its eventual answer or estimate whether continuing will succeed. For a transmitted cache, the relevant target may be a fact that should agree with the sender's public statement. In each case, the task defines both the possible answers and an observable criterion for evaluating them. This structure makes direct prediction over candidate answers a natural alternative to generating a textual interpretation. It also permits uncertainty to be expressed as a set of plausible answers or a calibrated alarm threshold.

We propose the Calibrated Latent Reader, \clr{}, a compact framework for predicting reasoning outcomes and recovering message content. Its central design is to evaluate candidate answers jointly with the observed representation, supporting answer prediction, correctness estimation and reference-value membership in one architecture. Source adapters enable transfer across models, while conformal calibration produces answer sets and message alarms. For communication, the alarms compare facts read from a cache with facts implied by the sender's public commitment.

Our contributions are as follows.
\begin{itemize}
\item We introduce \clr{}, a compact reader with explicit state--candidate interactions, unbiased continuation-value labels and conformal answer sets. It improves answer and membership prediction over probes by 4--18 points, approaching a 1.7B verbalizer's GPT-2 accuracy using 437--648 times fewer inference FLOPs.
\item We demonstrate adapter-only transfer between CODI sources. With 100--300 target labels, it outperforms MLP probes and readers trained from scratch on both tasks in both directions.
\item We develop a factual-decomposition audit with conformal false-alarm control under exchangeability. With \clr{}, it detects 93.1\% of KV-message substitutions at nominal 5\% FPR in the synthetic Qwen3-4B relay.
\item We establish two limits through theory and experiments: between-problem prediction need not support sibling ranking, and linear message compression leaves a null space invisible to the auditor.
\end{itemize}

\section{Related Work}
\label{sec:related}

\paragraph{Latent reasoning and monitoring.} Knowledge distillation compresses textual reasoning into continuous representations~\cite{deng2023implicit,shen2025codi}. Coconut propagates hidden states between reasoning steps, while Huginn scales computation through recurrent depth~\cite{hao2024training,geiping2026scaling}. Mirage and vision-aligned latent reasoning incorporate visual features into latent trajectories~\cite{yang2026machine,jeon2026vision}; broader approaches are reviewed by Zhu et al.~\cite{zhu2025survey}. Reading those representations does not establish that the model uses the decoded information causally~\cite{qiu2026does}. Related concerns arise in textual reasoning, where explanations can omit influential factors and monitoring can degrade under optimization~\cite{turpin2023language,baker2025monitoring,korbak2025chain}. Ulterior Motives studies misalignment probes for continuous thoughts~\cite{ramjee2026ulterior}. The targets here are explicitly defined predictions about source behavior or task annotations.

\paragraph{Reading model activations.} Linear probes and their nonlinear extensions predict specified properties of a representation~\cite{alain2016understanding,belinkov2022probing}; control tasks help separate representational information from probe capacity~\cite{hewitt2019controls}. The tuned lens learns intermediate vocabulary readouts~\cite{belrose2023tuned}. Patchscopes and SelfIE use activation patching for natural-language interpretation~\cite{ghandeharioun2024patchscopes,chen2024selfie}, and LatentQA learns a decoder from activation--question--answer pairs~\cite{pan2026latentqa}. Universal Activation Verbalizer extends this approach across models through adapters~\cite{zhao2026universal}. Verbalizer knowledge can confound such interpretations~\cite{li2025activation}; Faithful Activation Verbalization improves recoverability through explicit text reconstruction~\cite{zhao2026faithful}. \clr{} instead scores task-defined answers directly. Its evaluation compares prediction accuracy, uncertainty and cost, with controls that remove or replace the latent evidence.

\paragraph{Verification and inference-time selection.} Process reward models evaluate written reasoning steps~\cite{lightman2024let,wang2024math}. LTO learns correctness signals from latent thoughts~\cite{du2026latent}, whereas HSRM and CLUE verify explicit reasoning through hidden states~\cite{li2026hsrm,liang2025clue}. Such estimates can guide inference-time search~\cite{snell2024scaling}, but aggregate discrimination does not establish useful ranking among candidates for the same problem. Our experiments measure both quantities. Narrow Decision Models~\cite{antunes2026narrow} provide the option-isolation design used by \clr{}; the present study applies it to latent observations, with behavioral supervision, source transfer and content auditing.

\paragraph{Predictive uncertainty.} Temperature scaling adjusts class probabilities~\cite{guo2017calibration}. Conformal prediction constructs answer sets with coverage under exchangeability~\cite{vovk2005algorithmic,angelopoulos2022gentleintroductionconformalprediction,romano2020adaptive}, and weighted variants address specified covariate shifts~\cite{tibshirani2019conformal}. Semantic entropy estimates uncertainty over equivalent language generations~\cite{kuhn2023semantic}; activation-oracle work studies confidence from candidate scoring~\cite{torrielli2026confidence}. Finite answer spaces allow \clr{} predictions to be calibrated directly against observed outcomes, separately for each task and reasoning step. Distribution shifts are evaluated empirically rather than covered by the exchangeability guarantee.

\paragraph{Latent communication.} LatentMAS and Cache-to-Cache relay continuous representations between agents~\cite{zou2025latent,fu2026cache}. Subsequent work examines cache integrity, latent attacks and sensitive-information leakage~\cite{brito2026latent,wang2026out,asif2026lcguard}, as well as the causal contribution of the transmitted channel~\cite{cheng2026does,zhang2026latent}. Consilience calibrates communication actions in a dialogue~\cite{babu2026consilience}. Our audit instead compares facts read from a message with facts implied by a public commitment. Its attack analysis characterizes the message directions that the compressed observation cannot expose, distinguishing content verification from transport integrity.

\section{Calibrated Latent Reader}
\label{sec:method}

\subsection{Problem formulation}
\label{sec:reader-task}
The source model $M_s$ is fixed throughout reader training. Its observation $z$ is either a reasoning prefix $z_{0:t}$, with $z_i\in\mathbb R^{d_s}$, or a compressed message $f(m)$ derived from a key--value cache $m=\{(K_l,V_l)\}_{l=1}^{L}$. The source index $s$ identifies the representation space. A task specification $q$ determines a target type $\kappa(q)$ and an answer space $\mathcal A_q$. The reader estimates
\begin{equation}
R_\theta(z,q,s)=p_\theta(\cdot\mid z,q,s)\in\Delta(\mathcal A_q),
\label{eq:reader}
\end{equation}
where $\Delta(\mathcal A_q)$ denotes distributions over possible answers. The inputs contain the observed states and task specification, excluding the original problem, private sender inputs and future states.

The task and output format have separate roles. A final-answer task uses a choice distribution over candidates; correctness and membership tasks use $\mathcal A_q=\{0,1\}$. A value estimate is the probability of outcome 1. For type-specific datasets $\mathcal D_\kappa$ of observed examples $(z,q,s,y)$, the training objective is
\begin{equation}
\mathcal L(\theta)=\sum_\kappa\frac{1}{|\mathcal D_\kappa|}
\sum_{(z,q,s,y)\in\mathcal D_\kappa}-\log p_\theta(y\mid z,q,s).
\label{eq:loss}
\end{equation}
This weights task types equally and uses Bernoulli log loss for value prediction. The parameters $\theta$ include the adapters, transformer and prediction heads. Calibration is performed after training.

\subsection{Task-conditioned representation reading}
\label{sec:reader-architecture}
\begin{figure*}[t]
\centering
\includegraphics[width=\linewidth]{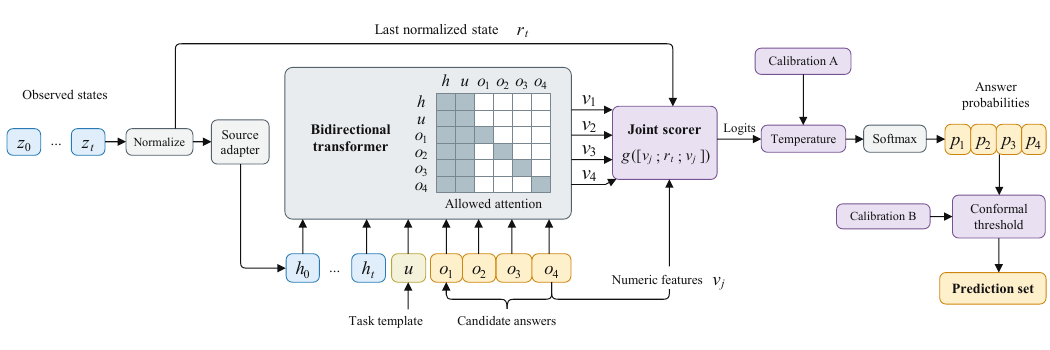}
\caption{Reader architecture for choice prediction. The masked transformer evaluates each candidate against the latent and task context. The scorer combines its marker $v_j$, the last normalized state $r_t$ and numeric features $\nu_j$. Probability adjustment and conformal set construction use separate calibration sets.}
\label{fig:arch}
\end{figure*}

\Cref{fig:arch} gives the reader architecture. States are standardized using source training statistics. A two-layer adapter maps a normalized state $r_i$ to the reader width $d$, giving $h_i=A_s(r_i)+e_i^{\mathrm{step}}$, where $e_i^{\mathrm{step}}$ represents its position in the reasoning prefix. Joint training across sources uses separate input alignments and a source token. For transfer, a linear alignment maps the new representation into a trained reader whose adapter, transformer and heads remain fixed.

The prediction task is encoded by template tokens $u$ and a type embedding. Each candidate has a token block $o_j$ and eight numeric features $\nu_j$. A bidirectional transformer processes $[h_{0:t};u;o_1;\ldots;o_n]$. Following Narrow Decision Models~\cite{antunes2026narrow}, attention is restricted so that the latent and task tokens form a shared context, while each candidate attends only to that context and its own block. Thus each candidate is evaluated against the same evidence without attending to the other candidate blocks.

The final candidate marker $v_j$ is scored together with the last normalized input state $r_t$ and its numeric features:
\begin{equation}
\ell_j=g_\theta([v_j;r_t;\nu_j]),\qquad
p_\theta(a_j\mid z,q,s)=\frac{\exp\ell_j}{\sum_b\exp\ell_b}.
\label{eq:choice}
\end{equation}
The brackets concatenate features, and $g_\theta$ is a shared nonlinear head. The direct state input supplies an explicit interaction between the observed representation and each candidate. Binary and value heads use a sigmoid. For correctness tasks, the task marker replaces $v_j$ and the numeric features are zero; membership tasks retain the candidate representation.

\subsection{Supervision from behavior and reference solutions}
\label{sec:labels}
Four tasks provide the reasoning labels. Q1 estimates final correctness after continued computation; Q2 predicts correctness if the source stops and decodes immediately. Q3 identifies the source's final answer among four candidates. Its distractors come from sibling continuations, the reference answer and other problems. Q4 predicts whether a candidate value appears in the reference solution. Negative values are drawn from other reference solutions to match the positive value marginal. Each task has six language templates, of which two are reserved for evaluating generalization.

For Q1, the target depends on the continuation policy. Policy $\pi_\sigma$ perturbs each subsequent latent input and then decodes greedily:
\begin{equation}
\tilde z_i=z_i+\frac{\sigma\|z_i\|}{\sqrt{d_s}}\epsilon_i,
\quad \epsilon_i\sim\mathcal N(0,I),\quad i>t.
\label{eq:policy}
\end{equation}
The target value is $V_\sigma(z_{0:t})=\Pr_{\pi_\sigma}[\text{final answer correct}\mid z_{0:t}]$. A trajectory perturbed after $t_0$ produces one binary outcome $y$. That outcome labels the clean state at $t_0$ and the later perturbed states, because their remaining transitions follow $\pi_\sigma$. Earlier clean prefixes would correspond to a different continuation policy.
\begin{proposition}[Unbiased value labels]
\label{prop:labels}
For every labeled state, $\mathbb E[y\mid z_{0:t}]=V_\sigma(z_{0:t})$, and the population Q1 log loss in \cref{eq:loss} is minimized over measurable readers by $V_\sigma$. Labeling an earlier clean prefix $t<t_0$ would instead target a policy that delays perturbations until $t_0$.
\end{proposition}
The rollout and noise settings are given in \cref{sec:setup}.

\subsection{Calibrated predictions}
\label{sec:calibration}
Two calibration sets separate probability adjustment from prediction-set construction. On $\mathrm{cal}_A$, minimizing NLL gives a positive temperature $T_\kappa$ for each task type and replaces its logits with $\ell/T_\kappa$. In the following, $p_\theta$ denotes the resulting fixed distribution.

For a fixed source, Mondrian calibration groups examples by task and step, $g=(\kappa,t)$, using one query per problem within each group. Define the nonconformity score $S(z,q,a)=1-p_\theta(a\mid z,q,s)$. From the $n_g$ true-answer scores in $\mathrm{cal}_B$, take the $\lceil(n_g+1)(1-\alpha)\rceil$-th order statistic as $\hat s_g$. The answer set is
\begin{equation}
C(z,q)=\{a\in\mathcal A_q:S(z,q,a)\le\hat s_g\}.
\label{eq:cp}
\end{equation}
If the required rank exceeds $n_g$, the threshold is $+\infty$ and all answers are retained. Q1 coverage applies to a future binary outcome, rather than an interval for the mean $V_\sigma$. The guarantee requires exchangeability within each group.

\subsection{Content auditing of cache messages}
\label{sec:audit-method}
A sender observes public input $x_{\mathrm{pub}}$ and private input $x_{\mathrm{priv}}$, sends cache $m$, and publishes commitment $c$. The receiver sees the public input and cache. To read a manageable representation, a probe scan on training data selects four layers $\mathcal L$. Token groups $\mathcal S_r$ pool log-record segments and the public prefix, while keeping the final eight valid tokens as singleton groups. A fixed projection $P\in\mathbb R^{512\times D}$ with orthonormal rows gives
\begin{equation}
f(m)_{l,r}=\frac{P}{|\mathcal S_r|}
\sum_{j\in\mathcal S_r}[K_{l,j};V_{l,j}],\quad l\in\mathcal L.
\label{eq:feat}
\end{equation}
Here $D$ is the concatenated key--value width. This linear compression supplies the message adapter.

The reading tasks cover fact membership in A1, conclusion support in A2, commitment agreement in A3 and private-value recovery in A4. Auditing decomposes a commitment into $k$ factual queries. Their expected answers $\hat a_i=e(q_i,c,x_{\mathrm{pub}})$ follow from public information alone. Agreement with the message readout is measured by
\begin{equation}
\Gamma(m,c)=\frac1k\sum_{i=1}^k
\log p_\theta(\hat a_i\mid f(m),q_i,s).
\label{eq:audit}
\end{equation}
An alarm is raised when $\Gamma(m,c)<\tau_\alpha$. With $n$ clean messages in $\mathrm{cal}_B$, $\tau_\alpha$ is the $\lfloor\alpha(n+1)\rfloor$-th smallest score, taking $-\infty$ at rank zero. Direct support and agreement predictions are evaluated separately from this factual decomposition.

\begin{proposition}[Coverage and false alarms]
\label{prop:coverage}
(a) For exchangeable calibration and test pairs within $g$, \cref{eq:cp} covers with probability at least $1-\alpha$, and at most $1-\alpha+1/(n_g+1)$ without ties. (b) For exchangeable clean message scores, $\Pr[\mathrm{alarm}]\le\alpha$. (c) Conditional on a substituted message, independent question log-probabilities in $[-B,0]$ with mean $\mu_1<\tau$ give miss probability at most $\exp[-2k(\tau-\mu_1)^2/B^2]$ for fixed threshold $\tau$.
\end{proposition}
The coverage claims use split conformal prediction~\cite{vovk2005algorithmic}; the detection bound follows Hoeffding's inequality. Proofs are in \cref{app:proofs}. Repeated queries about one fact can be correlated, so adding queries need not improve detection. The guarantee controls false alarms on exchangeable clean messages; adaptive attacks are evaluated in \cref{sec:robust}.

\section{Experiments}
\label{sec:experiments}
\label{sec:results}

\begin{table*}[t]
\centering
\caption{Prediction from CODI states on GSM8K. Each probe answers one question type; \clr{} answers all four. Joint training shares the reader across both sources. Bold and underlining mark the best and second-best results; shaded rows use \clr{}.}
\label{tab:main}
\small
\setlength{\tabcolsep}{3pt}
\begin{tabular}{l c cccc c cccc}
\toprule
& & \multicolumn{4}{c}{CODI-GPT-2} && \multicolumn{4}{c}{CODI-LLaMA-1B}\\
\cmidrule{3-6}\cmidrule{8-11}
Reader & Types & \shortstack{Value\\NLL$\downarrow$} & \shortstack{Early exit\\AUROC$\uparrow$} & \shortstack{Answer\\Acc. (\%)$\uparrow$} & \shortstack{Membership\\Acc. (\%)$\uparrow$} && \shortstack{Value\\NLL$\downarrow$} & \shortstack{Early exit\\AUROC$\uparrow$} & \shortstack{Answer\\Acc. (\%)$\uparrow$} & \shortstack{Membership\\Acc. (\%)$\uparrow$}\\
\midrule
\multicolumn{11}{l}{\textit{Probes}}\\
Question only & one & 0.696 & 0.508 & 35.9 & 49.8 && 0.692 & 0.509 & 39.1 & 50.1\\
Linear probe~\cite{alain2016understanding} & one & 0.553 & 0.836 & 53.4 & 50.1 && 0.615 & 0.764 & 54.0 & 50.1\\
MLP probe & one & 0.523 & 0.846 & 58.6 & 68.3 && 0.612 & \second{0.851} & 58.9 & 62.7\\
MLP probe, prefix & one & \second{0.515} & \textbf{0.857} & 61.8 & 68.4 && 0.606 & \textbf{0.852} & 59.3 & 60.3\\
Attention probe & one & 0.574 & \second{0.853} & 65.4 & 72.0 && 0.632 & 0.824 & 57.8 & 66.6\\
\midrule
\multicolumn{11}{l}{\textit{Prior latent readers}}\\
LTO reward model~\cite{du2026latent} & Q1, Q2 & \textbf{0.507} & 0.845 & -- & -- && \second{0.551} & 0.806 & -- & --\\
HSRM~\cite{li2026hsrm} & Q1, Q2 & 0.559 & 0.834 & -- & -- && 0.625 & 0.834 & -- & --\\
CLUE~\cite{liang2025clue} & Q1, Q2 & 0.648 & 0.575 & -- & -- && \textbf{0.545} & 0.793 & -- & --\\
Patchscope, zero-shot~\cite{ghandeharioun2024patchscopes} & all & 0.658 & 0.450 & 24.5 & 50.4 && -- & -- & -- & --\\
LatentQA, 1.7B~\cite{pan2026latentqa} & Q3, Q4 & -- & -- & \textbf{76.8} & \second{76.3} && -- & -- & -- & --\\
\midrule
\rowcolor{oursrow}\clr{}, single step & all & 0.564 & 0.819 & 73.8 & 74.8 && 0.627 & 0.813 & \second{74.7} & \second{75.5}\\
\rowcolor{oursrow}\clr{} & all & 0.548 & 0.830 & 76.1 & 76.2 && 0.615 & 0.821 & 72.1 & 74.2\\
\rowcolor{oursrow}\clr{}, joint & all & 0.558 & 0.831 & \second{76.2} & \textbf{76.4} && 0.622 & 0.798 & \textbf{75.0} & \textbf{76.5}\\
\bottomrule
\end{tabular}
\end{table*}

\subsection{Experimental setup}
\label{sec:setup}

\paragraph{Sources and data.} We use released CODI checkpoints on GPT-2 and LLaMA-3.2-1B-Instruct, each with six latent steps. Single-precision inference at batch size one agrees with official GSM8K~\cite{cobbe2021training} predictions on 95.3\% and 98.9\% of problems. Readers train on 20k and 6.8k GSM8k-Aug~\cite{deng2023implicit} problems previously seen by the respective sources. Evaluation removes 13-gram overlap with source training, using 500 calibration and 1005 test problems. Final-step early-exit accuracy is 82\% on source-training problems versus 42\% on test problems in \cref{fig:search}a. We also evaluate SVAMP~\cite{patel2021nlp} and ASDiv~\cite{miao2020diverse}, where GPT-2 source accuracy is 43\% and 55\%.

\paragraph{Continuation targets.} Q1 uses 16 rollouts per clean state. We choose noise to produce non-degenerate values while losing fewer than five source-accuracy points, giving $\sigma=1.5$ for GPT-2 and $\sigma=2$ for LLaMA.

\paragraph{Baselines.} We compare a question-only reader using zeroed states, a linear probe, two-layer MLPs using the current state or full prefix, and a single-layer attention probe over the prefix. Probes train separately per task; attention receives at least the MLP's exposure and parameter budget. LTO~\cite{du2026latent} predicts scalar correctness rewards. Patchscope~\cite{ghandeharioun2024patchscopes} patches states into the source model without training. Our LatentQA verbalizer~\cite{pan2026latentqa} uses a two-layer projection into frozen Qwen3-1.7B~\cite{yang2025qwen3}, with rank-16 LoRA trained on the same reading data. HSRM~\cite{li2026hsrm} uses a two-layer transformer; CLUE~\cite{liang2025clue} compares $z_t-z_0$ with success and failure centroids.

\paragraph{Evaluation.} Readers are selected by validation NLL. Results average three seeds, except deterministic CLUE, with equal step weights, clean states and training templates. Q1 measures continuation-outcome NLL, Q2 early-exit AUROC, Q3 answer accuracy and Q4 reference-value membership accuracy; chance is 25\% for Q3 and 50\% for Q4. Readers share the calibration procedure. The 95\% intervals bootstrap whole problems or audit episode groups. FLOPs count matrix and attention operations at two per multiply-add; timing excludes feature preparation, tokenization and transfers.

\subsection{Prediction from latent reasoning states}
\label{sec:exp-read}

To evaluate latent reading, \cref{tab:main} compares \clr{} with probes, reward models and language-model readers. \clr{} improves final-answer and membership accuracy over the MLP and attention probes by 4--18 points across sources, with paired intervals excluding zero. On membership, the linear probe stays at chance versus 76.2\% and 74.2\% for \clr{}, supporting nonlinear state--candidate relations. On correctness prediction, \clr{} is within 0.071 NLL and 0.031 AUROC of the best specialist while sharing one model across four types.

To distinguish state information from candidate priors, \cref{tab:main} compares input controls and prefix access. Zeroed or permuted states predict at chance apart from the answer prior in the options. Providing the MLP with the full prefix raises answer accuracy only from 58.6\% to 61.8\% on GPT-2 and from 58.9\% to 59.3\% on LLaMA. Current-state \clr{} reaches 73.8\% and 74.7\%, preserving its advantage with matched state access. Zero-shot Patchscope remains near chance on these tasks.

\subsection{Ablation and efficiency analysis}
\label{sec:exp-ablation}
\begin{table*}[t]
\centering
\caption{Reader design ablations on the two CODI sources. Each row changes one component of \clr{}; separate readers use matched training exposure. Answer-set size is measured at nominal 90\% coverage. Bold denotes the best result in each column. $^\dagger$/$^\ddagger$ indicate worse/better results with a paired 95\% difference interval excluding zero. Dashes indicate configurations not evaluated on that source.}
\label{tab:ablation}
\small
\setlength{\tabcolsep}{2.5pt}
\begin{tabular}{lccccc@{\hspace{9pt}}ccccc}
\toprule
& \multicolumn{5}{c}{CODI-GPT-2} & \multicolumn{5}{c}{CODI-LLaMA-1B}\\
\cmidrule(lr){2-6}\cmidrule(lr){7-11}
Variant & \shortstack{Value\\NLL$\downarrow$} & \shortstack{Early exit\\AUROC$\uparrow$} & \shortstack{Answer\\Acc. (\%)$\uparrow$} & \shortstack{Membership\\Acc. (\%)$\uparrow$} & \shortstack{Answer\\Set size$\downarrow$} & \shortstack{Value\\NLL$\downarrow$} & \shortstack{Early exit\\AUROC$\uparrow$} & \shortstack{Answer\\Acc. (\%)$\uparrow$} & \shortstack{Membership\\Acc. (\%)$\uparrow$} & \shortstack{Answer\\Set size$\downarrow$}\\
\midrule
\rowcolor{oursrow}\clr{} & \textbf{0.548} & 0.830 & 76.1 & 76.2 & 1.49 & \textbf{0.615} & \textbf{0.821} & 72.1 & 74.2 & 1.64\\
\midrule
\multicolumn{11}{l}{\textit{Candidate conditioning}}\\
w/o block mask & 0.568$^\dagger$ & 0.822$^\dagger$ & 75.1$^\dagger$ & 75.4 & 1.54$^\dagger$ & 0.619$^\dagger$ & 0.811$^\dagger$ & 64.0$^\dagger$ & 71.9$^\dagger$ & 1.91$^\dagger$\\
Dot-product head & 0.584$^\dagger$ & 0.819$^\dagger$ & 72.9$^\dagger$ & 74.3$^\dagger$ & 1.58$^\dagger$ & 0.627$^\dagger$ & 0.786$^\dagger$ & 68.4$^\dagger$ & 73.3 & 1.78$^\dagger$\\
w/o numeric features & 0.549 & 0.834 & 75.0$^\dagger$ & 75.4 & 1.49 & -- & -- & -- & -- & --\\
\midrule
\multicolumn{11}{l}{\textit{Representation and training}}\\
Late fusion & 0.557 & \textbf{0.838} & 74.0$^\dagger$ & \textbf{76.7} & 1.54$^\dagger$ & 0.628$^\dagger$ & 0.819 & \textbf{75.7}$^\ddagger$ & \textbf{76.4}$^\ddagger$ & \textbf{1.45}$^\ddagger$\\
Pretrained encoder & 0.564$^\dagger$ & 0.830 & 76.2 & 75.0$^\dagger$ & 1.52$^\dagger$ & -- & -- & -- & -- & --\\
w/o state history & 0.564$^\dagger$ & 0.819$^\dagger$ & 73.8$^\dagger$ & 74.8$^\dagger$ & 1.61$^\dagger$ & 0.627$^\dagger$ & 0.813 & 74.7$^\ddagger$ & 75.5$^\ddagger$ & 1.50$^\ddagger$\\
w/o task weighting & 0.575$^\dagger$ & 0.829 & 75.9 & 76.6 & 1.51$^\dagger$ & -- & -- & -- & -- & --\\
w/o task sharing & 0.557 & 0.823 & \textbf{76.7} & 73.7$^\dagger$ & \textbf{1.48}$^\ddagger$ & 0.634$^\dagger$ & 0.818 & 72.6 & 71.4$^\dagger$ & 1.64\\
\bottomrule
\end{tabular}
\end{table*}

\paragraph{Candidate conditioning.} To isolate candidate interactions, \cref{tab:ablation} removes option isolation or replaces joint scoring with a dot product. Removing isolation costs 1.0 answer-accuracy points on GPT-2 and 8.1 on LLaMA, enlarging sets from 1.49 to 1.54 and 1.64 to 1.91 options. Dot-product scoring loses 3.2 and 3.7 accuracy points. Both components therefore contribute across sources. Removing numeric features costs a further 1.1 points on GPT-2 answer prediction.

\paragraph{Representation and supervision.} To assess shared supervision, \cref{tab:ablation} compares separate readers at matched exposure. Sharing improves membership accuracy by 2.5 points on GPT-2 and 2.8 on LLaMA, with no detectable change in answer accuracy. Prefix length and fusion depth are source dependent: the full prefix and early fusion help GPT-2, whereas current-state input and late fusion give higher LLaMA answer accuracy. A pretrained encoder uses seven times the trainable parameters without improving GPT-2 answer accuracy.

\paragraph{Accuracy and computation.} To assess the cost of direct prediction, \cref{fig:teaser} compares answer accuracy with inference FLOPs, and \cref{tab:main} includes the 1.7B LatentQA verbalizer. The GPT-2 accuracy gaps are 0.7 points for answers and 0.1 for membership. Both satisfy paired equivalence tests within a prespecified $\pm2$-point margin. \clr{} uses 437--648 times fewer FLOPs, with 19--80 times lower measured latency across batch sizes one and 64. At batch size one, a query costs 0.6--0.9 cached CODI-GPT-2 steps. Direct scoring thus provides equivalent measured accuracy with less computation.

\begin{table}[t]
\centering
\caption{Calibrated answer sets on CODI-GPT-2 at nominal 90\% coverage.}
\label{tab:calibration}
\small
\setlength{\tabcolsep}{4pt}
\begin{tabular}{llcc}
\toprule
Task & Reader & Coverage (\%) & Set size$\downarrow$\\
\midrule
\multirow{2}{*}{Final answer} & MLP probe & 92.0 & 2.16\\
& \clr{} & 91.6 & 1.49\\
\midrule
\multirow{2}{*}{Membership} & MLP probe & 90.1 & 1.53\\
& \clr{} & 89.7 & 1.35\\
\bottomrule
\end{tabular}
\end{table}

\paragraph{Calibrated answer sets.} To assess calibrated uncertainty, \cref{tab:calibration} compares coverage and set size at the nominal 90\% level. \clr{} reduces the mean answer set from 2.16 to 1.49 options relative to the MLP, while empirical coverage is 91.6\% versus 92.0\%. Membership sets shrink from 1.53 to 1.35, with coverage of 89.7\% versus 90.1\%. Paired set-size difference intervals exclude zero for both tasks. Compared with the verbalizer, \clr{} retains fewer than 0.02 additional options on average. The accuracy gains thus yield more selective sets at similar coverage.

\subsection{Transfer and generalization}
\label{sec:exp-cross}

\begin{figure}[t]
\centering
\includegraphics[width=\linewidth]{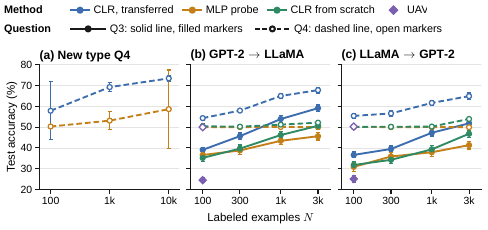}
\caption{Transfer with few labels. Solid lines show Q3 and dashed lines Q4. (a) Fine-tuning on $N$ labels of the unseen type Q4. (b, c) Adapter-only transfer between CODI sources; the UAV baseline is evaluated at $N=100$. Bars are 95\% intervals.}
\label{fig:transfer}
\end{figure}

\paragraph{Transfer to a new question type.} To test reuse across tasks, \cref{fig:transfer}a evaluates fine-tuning on membership questions after training on the other three types. The transferred reader exceeds an MLP trained from scratch at every budget from 100 to 10k labels. This complements the benefit of shared supervision in \cref{tab:ablation}.

\paragraph{Transfer across sources.} To evaluate reuse across source models, \cref{fig:transfer}b,c fixes the donor reader and trains only an input adapter. At 100 or 300 labels, transfer beats an MLP and \clr{} trained from scratch on both tasks in both directions. At 100 labels it also exceeds UAV~\cite{zhao2026universal}, a linear map into frozen Qwen3-1.7B. Transferred membership prediction rises above chance while training from scratch remains near it. Joint-source training in \cref{tab:main} further improves LLaMA answer and membership accuracy and reduces its answer-set size, without a detectable accuracy change on GPT-2.

\paragraph{Coverage under distribution shift.} To assess calibration under shift, we evaluate source transfer in \cref{fig:transfer}, GSM8K-to-SVAMP/ASDiv transfer and early-to-late steps. Empirical coverage remains within five points of nominal 90\% in all 32 settings, without density-ratio correction. Coverage under these shifts is empirical; \cref{prop:coverage} requires exchangeability within calibration strata.

\subsection{Content auditing of latent messages}
\label{sec:exp-audit}

\paragraph{Communication setting.} We use a frozen Qwen3-4B sender and receiver connected through a KV-cache relay following LatentMAS~\cite{zou2025latent}. Synthetic episodes supply known private facts and counterfactual pairs differing in one fact. Public commitments use three paraphrase templates shared across message classes. Training, validation, $\mathrm{cal}_A$, $\mathrm{cal}_B$ and test contain 8000, 1000, 500, 1000 and 2000 episode groups, with each pair confined to one split. Changing a private fact changes the receiver's target-answer probability by 13.8 points, confirming that the relay carries behaviorally relevant information.

\begin{table}[t]
\centering
\caption{Content auditing on 2000 synthetic KV-message groups. Fact and private-value columns give reading accuracy in \%. Detection is measured at thresholds calibrated for 5\% or 1\% false alarms. Factual decomposition uses 16 sub-questions with the listed reader.}
\label{tab:audit}
\small
\setlength{\tabcolsep}{3pt}
\begin{tabular}{lccccc}
\toprule
& & & \multicolumn{2}{c}{Detection (\%)$\uparrow$} & \\
\cmidrule(lr){4-5}
Auditor & Fact & \shortstack{Private\\value} & @5\% & @1\% & AUROC\\
\midrule
\multicolumn{6}{l}{\textit{Direct message audits}}\\
KV norm & -- & -- & 6.5 & 0.9 & 0.500\\
Text only & -- & -- & 0.0 & 0.0 & 0.533\\
Clean/CF classifier & -- & -- & 4.9 & -- & 0.500\\
\midrule
\multicolumn{6}{l}{\textit{Factual decomposition}}\\
Question only & 50.0 & 25.3 & 4.3 & -- & 0.500\\
Linear probe & -- & -- & 21.8 & -- & 0.733\\
MLP probe & 68.3 & 45.5 & 24.0 & 13.3 & 0.735\\
Attention probe & 71.0 & 98.0 & \textbf{97.9} & \textbf{92.8} & \textbf{0.994}\\
\rowcolor{oursrow}\clr{} & \textbf{99.5} & \textbf{99.3} & 93.1 & 83.9 & 0.986\\
\bottomrule
\end{tabular}
\end{table}

\paragraph{Substitution detection.} To evaluate factual decomposition, \cref{tab:audit} compares it with direct message scores. The \clr{} audit detects 93.1\% of substitutions at nominal 5\% false alarms and 83.9\% at nominal 1\%, with measured false-alarm rates of 5.07\% and 0.93\%. Direct support and consistency prediction, a clean/counterfactual classifier, and norm or text scores provide little separation. The attention probe reaches 97.9\% detection at the 5\% threshold despite lower fact accuracy, 71.0\% versus \clr{}'s 99.5\%. Detection thus depends on the relation to the commitment as well as reading accuracy.

\begin{figure}[t]
\centering
\includegraphics[width=\linewidth]{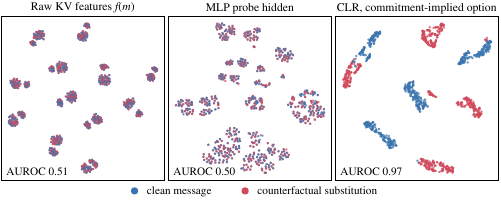}
\caption{t-SNE of 600 clean and 600 counterfactual test messages. Raw KV features and the MLP probe's hidden layer group messages by episode content and mix the two classes; \clr{}'s representation of the option implied by the commitment separates them. AUROC is five-fold logistic regression on the plotted vectors.}
\label{fig:qual-audit}
\end{figure}

\paragraph{Evidence in the learned representation.} To examine the audit representation, \cref{fig:qual-audit} compares raw KV features, MLP states and \clr{}'s representation of the commitment-implied option. The first two group by episode content; the last separates substitutions. A segment intervention localizes private-value evidence: reading its log segment alone gives 99.9\% accuracy, while removing that segment reduces accuracy to 2.5\%. Commitment agreement can hold even when the sender's final answer is wrong.

\subsection{Analysis of downstream decisions}
\label{sec:analysis}

\begin{figure}[t]
\centering
\includegraphics[width=\linewidth]{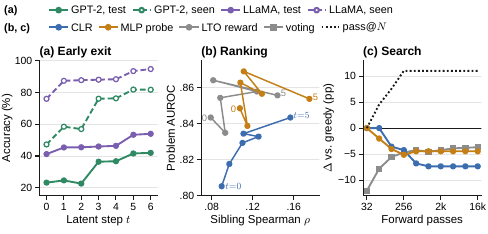}
\caption{(a) Early-exit accuracy by latent step. Solid lines show unseen test problems; dashed lines show problems seen in CODI's training. (b) Between-problem AUROC against within-sibling Spearman $\rho$, one point per step. (c) Search accuracy minus greedy at matched compute; pass@$N$ is an upper reference.}
\label{fig:search}
\end{figure}

\paragraph{Selecting latent continuations.}
\label{sec:search}
To assess candidate selection, \cref{fig:search} compares aggregate prediction, sibling ranking and search accuracy. Search ranks $N$ perturbed children and the unperturbed child from \cref{eq:policy}, either at the final step or throughout a beam. We compare the MLP, LTO, \clr{}, and \clr{} with a ranking loss or pretrained encoder over 90 configurations on 300 CODI-GPT-2 validation problems, using $N\le16$ and beam width at most four. Compute includes source forward passes and reader cost.

\Cref{fig:search}c shows a best search gain of $-0.9$ points over 42.3\% greedy accuracy, with 95\% interval $[-4.7,2.9]$, despite 53.3\% pass@16. To distinguish problem prediction from sibling ranking, we decompose value into a problem mean and residual,
\begin{equation}
V_\sigma(z)=\mu(x)+\nu(z),\qquad \mu(x)=\mathbb{E}[V_\sigma(z)\mid x],
\label{eq:decomp}
\end{equation}
where $x$ is the problem. All siblings share $\mu(x)$, so their ranking depends on $\nu$.
\begin{proposition}[Between-problem ranking does not imply sibling ranking]
\label{prop:within}
A reader $r(z)=\phi(\mu(x))$ with $\phi$ increasing attains the largest between-problem AUROC of any function of $x$, yet it is constant across siblings, and selecting $\arg\max_j r(z^{(j)})$ with ties broken at random gains $\mathbb{E}[\bar\nu]-\mathbb{E}[\nu(z^{(0)})]$ over the noise-free child, where $\bar\nu$ is the average residual of the candidate set. If $\sup_z|r(z)-V_\sigma(z)|\le\varepsilon$, best-of-$N$ selection loses at most $2\varepsilon$ against the best candidate.
\end{proposition}

\Cref{fig:search}b isolates this source of variation: across-problem AUROC reaches 0.82--0.86, whereas correlation with sibling Monte Carlo values is only Spearman $\rho\approx0.1$. Selection is slightly worse than retaining the unperturbed child, and neither a sibling-ranking loss nor smaller perturbations removes the gap. \Cref{prop:within} identifies the relevant accuracy requirement for search: small value error among the continuations of each problem.

\paragraph{Auditing adaptive cache perturbations.}
\label{sec:robust}
To assess adaptive changes to compressed observations, \cref{tab:robust} compares message and receiver-state auditors. For receiver distribution $p_{\mathrm{rec}}$, an attack seeks a counterfactual answer $y_{\mathrm{cf}}$ while passing the audit:
\begin{equation}
\begin{aligned}
\max_{\delta}\ &\log p_{\mathrm{rec}}(y_{\mathrm{cf}}\mid m+\delta)+\lambda\,\mathbb{E}_{q_{1:k}}\big[\Gamma(m+\delta,c)\big]\\
\text{s.t.}\ &\|\delta_l\|\le\epsilon\,\|m_l\|\quad\text{for every layer } l,
\end{aligned}
\label{eq:attack}
\end{equation}
using 200 PGD steps~\cite{madry2017towards} and EOT~\cite{athalye2018synthesizing} over sampled queries. Failure requires the target answer and acceptance at nominal 5\% false alarms. Restricted attacks use unread layers or the null space of $f$.
\begin{proposition}[Null-space blind spot]
\label{prop:nullspace}
If the auditor depends on $m$ only through a linear map $f$, then $\Gamma(m+\delta,c)=\Gamma(m,c)$ for every $\delta\in\ker f$. For $f$ in \cref{eq:feat}, $\dim\ker f\ge \dim m-512\,|\mathcal{L}|\,R$, where $R$ counts the pooled segments and the 8 kept tokens, leaving a high-dimensional subspace invisible to the auditor.
\end{proposition}

\begin{table}[t]
\centering
\caption{Audit failure under white-box cache perturbations, in \%. The largest relative norm budget is 0.3 for the relay receiver and 1 for the message-dependent receiver. Finite PGD measures an attack lower bound; its success need not increase with budget even though the true worst case cannot decrease.}
\label{tab:robust}
\small
\setlength{\tabcolsep}{4pt}
\begin{tabular}{llccc}
\toprule
& & \multicolumn{3}{c}{Relative norm budget $\epsilon$}\\
\cmidrule(lr){3-5}
Receiver & Auditor reads & 0.01 & 0.1 & Largest\\
\midrule
\multirow{2}{*}{Relay} & Message & 70.6 & 70.1 & 69.4\\
 & Receiver state & 71.5 & 70.9 & 70.5\\
\midrule
\multirow{2}{*}{Dependent} & Message & 96.5 & 96.0 & 95.3\\
 & Receiver state & 81.7 & 72.0 & 64.0\\
\bottomrule
\end{tabular}
\end{table}

\Cref{tab:robust} shows that 1\% relative-norm perturbations suffice for substantial joint receiver-change and audit-evasion rates. In the relay setting, the target answer is reached in every episode group while the message audit remains at its false-positive rate. Failure stays between 69\% and 72\% with 1--64 queries, secret compression seeds or access to every layer. Reading the receiver state reduces unread-layer failures from 70\% to 33\% and null-space failures from 69\% to 58\%, localizing part of the vulnerability to the observation itself.

To isolate receiver dependence, \cref{tab:robust} also uses a direct-lookup receiver with 93\% accuracy given the message and 27\% without it. Receiver-state reading reaches 99.5\% private-value accuracy, yet adaptive attacks evade both auditors; the median budget ratio between evading the audit and changing the receiver alone is one. The worst-case failure at $\epsilon=1$ is at least the 81.7\% already attained at $\epsilon=0.01$. \Cref{prop:nullspace} explains this distinction between calibrated content detection and adaptive integrity. Content auditing therefore complements authenticated transport~\cite{brito2026latent}.

\section{Conclusion}
\clr{} enables compact prediction of reasoning outcomes and message content from latent states. Across two CODI sources, state--candidate interactions improve answer and membership prediction. Shared supervision and source adapters support reuse across tasks and models with few labels. On GPT-2, reading accuracy is equivalent to a 1.7B verbalizer within the tested margin at substantially lower inference cost. Conformal calibration yields selective answer sets, and factual decomposition enables calibrated detection of message substitutions in synthetic communication.

\clearpage
\appendix

\section{Proofs}
\label{app:proofs}

Throughout, $z$ denotes a latent object, $x$ the problem or episode that produced it, and $y\in\{0,1\}$ an outcome. Expectations are over the randomness of the source model, of the policy $\pi_\sigma$ and of the data split.

\subsection{Proof of \texorpdfstring{\cref{prop:labels}}{Proposition \getrefnumber{prop:labels}}}
Let a training trajectory start perturbing at step $t_0$. By construction its latent inputs at steps $i\le t_0$ are the clean inputs, and its inputs at steps $i>t_0$ follow \cref{eq:policy} with fresh noise. Consider a state $z_{0:t}$ on this trajectory with $t\ge t_0$. Conditional on $z_{0:t}$, the remaining inputs $\tilde z_{t+1},\dots,\tilde z_T$ are generated by the same Markov kernel as under $\pi_\sigma$ started at $z_{0:t}$: the noise $\epsilon_{t+1},\dots,\epsilon_T$ is independent of everything up to step $t$, and the decoding rule is the same. The outcome $y$ is a deterministic function of the full input sequence and the problem. Hence
\[
\mathbb{E}[y\mid z_{0:t}]=\Pr_{\pi_\sigma}[\text{correct}\mid z_{0:t}]=V_\sigma(z_{0:t}).
\]
For the second claim, the Q1 term of \cref{eq:loss} is the empirical version of $\mathbb{E}\big[-y\log r(z)-(1-y)\log(1-r(z))\big]$ for a reader $r$ with values in $(0,1)$. Conditioning on $z$, the inner expectation equals $-V\log r-(1-V)\log(1-r)$ with $V=V_\sigma(z)$, which is strictly convex in $r$ and minimized at $r=V$. The minimizer over measurable readers is therefore $r=V_\sigma$ almost everywhere.

For a clean prefix $z_{0:t}$ with $t<t_0$, the inputs at steps $t+1,\dots,t_0$ are clean and only later inputs are perturbed. The same argument gives $\mathbb{E}[y\mid z_{0:t},t_0]=V_{\sigma}^{(t_0)}(z_{0:t})$, the value of the policy that follows the noise-free dynamics up to $t_0$ and $\pi_\sigma$ afterwards. This differs from $V_\sigma(z_{0:t})$ whenever perturbing steps $t+1,\dots,t_0$ changes the success probability, which is why such prefixes receive no Q1 label. \hfill$\square$

\subsection{Proof of \texorpdfstring{\cref{prop:coverage}}{Proposition \getrefnumber{prop:coverage}}}
\paragraph{Part (a).} Fix a stratum $g=(\kappa,t)$. Keeping one question instance per problem makes the calibration pairs $(z_i,q_i,a^\star_i)$, $i=1,\dots,n_g$, and the test pair $(z,q,a^\star)$ exchangeable by assumption. The reader $p_\theta$ and the temperature $T_\kappa$ were fitted on data disjoint from $\mathrm{cal}_B$ and the test set, so they are fixed functions here, and the scores $S_i=S(z_i,q_i,a^\star_i)$ and $S_{n_g+1}=S(z,q,a^\star)$ are exchangeable. Let $\hat k=\lceil(n_g+1)(1-\alpha)\rceil$. By exchangeability, the rank of $S_{n_g+1}$ among the $n_g+1$ scores is uniform on $\{1,\dots,n_g+1\}$ when ties are broken at random, so
\[
\Pr\big[S_{n_g+1}\le S_{(\hat k)}\big]\ge\frac{\hat k}{n_g+1}\ge1-\alpha .
\]
The event $S_{n_g+1}\le S_{(\hat k)}$ is the event $a^\star\in C(z,q)$ by \cref{eq:cp}. Without ties the probability equals $\hat k/(n_g+1)\le1-\alpha+1/(n_g+1)$. If $\hat k>n_g$ we set the threshold to $+\infty$, and the set is $\mathcal{A}_q$, which covers trivially; this is the case in our HiddenBench runs at $\alpha=0.01$.

\paragraph{Part (b).} Let $\Gamma_1,\dots,\Gamma_n$ be the scores of the clean calibration messages and $\Gamma_{n+1}$ the score of a clean test message, exchangeable by assumption, and let $j=\lfloor\alpha(n+1)\rfloor$. The alarm fires when $\Gamma_{n+1}<\Gamma_{(j)}$, which requires $\Gamma_{n+1}$ to be strictly smaller than at least $n+1-j$ of the other scores, so its rank among the $n+1$ scores is at most $j$. The probability of that is at most $j/(n+1)\le\alpha$. The question draws $q_{1:k}$ are part of each message's score and are drawn independently of the message, so they do not affect exchangeability.

\paragraph{Part (c).} Fix a substituted message $m$. Conditional on $m$, $\Gamma$ is the mean of $k$ independent variables in $[-B,0]$ with mean $\mu_1$. A miss is the event $\Gamma\ge\tau$, that is $\Gamma-\mu_1\ge\tau-\mu_1>0$. Hoeffding's inequality gives $\Pr[\Gamma-\mu_1\ge\tau-\mu_1]\le\exp\big(-2k(\tau-\mu_1)^2/B^2\big)$. The same argument bounds the false-alarm rate of the fixed threshold by $\exp(-2k(\mu_0-\tau)^2/B^2)$. When the $k$ answers are strongly correlated, for example because all questions are about the same private value, the effective number of independent terms is close to one and neither bound improves with $k$. \hfill$\square$

\subsection{Proof of \texorpdfstring{\cref{prop:within}}{Proposition \getrefnumber{prop:within}}}
\paragraph{Between-problem ranking.} Consider predicting $y$ from a function of $x$ alone. For a score $r=\phi(x)$, the AUROC is $\Pr[\phi(x)>\phi(x')\mid y=1,y'=0]+\tfrac12\Pr[\phi(x)=\phi(x')\mid y=1,y'=0]$ for independent pairs. Since $\Pr[y=1\mid x]=\mu(x)$, the likelihood ratio $p(x\mid y=1)/p(x\mid y=0)$ is an increasing function of $\mu(x)$, and by the Neyman--Pearson lemma every threshold rule on $\mu(x)$ has the largest true-positive rate at its false-positive rate among rules based on $x$. The ROC curve of $\mu$ therefore dominates that of every other function of $x$, and so does its AUROC. Increasing transformations $\phi$ leave the ROC curve unchanged.

\paragraph{Selection gain.} Siblings $z^{(0)},\dots,z^{(N)}$ share $x$, so $r(z^{(j)})=\phi(\mu(x))$ is the same for all $j$, and random tie-breaking selects an index $J$ uniform on $\{0,\dots,N\}$ independently of the candidates. Then
\[
\begin{aligned}
&\mathbb{E}[V_\sigma(z^{(J)})]-\mathbb{E}[V_\sigma(z^{(0)})] \\
&\quad=\mathbb{E}\Big[\frac{1}{N+1}\sum_{j}\nu(z^{(j)})\Big]-\mathbb{E}[\nu(z^{(0)})],
\end{aligned}
\]
because $\mu(x)$ cancels. In our data the perturbed children have lower value than the noise-free child on average, so this gain is negative.

\paragraph{Uniform error.} Let $J=\arg\max_j r(z^{(j)})$ and $J^\star=\arg\max_j V_\sigma(z^{(j)})$. If $|r-V_\sigma|\le\varepsilon$ everywhere, then $V_\sigma(z^{(J)})\ge r(z^{(J)})-\varepsilon\ge r(z^{(J^\star)})-\varepsilon\ge V_\sigma(z^{(J^\star)})-2\varepsilon$. Applying this at every step of a beam of width one gives the same bound per step. A reader with small error across problems can still have $|r-V_\sigma|$ large relative to the spread of $\nu$ within a problem, and the bound is informative when $\varepsilon$ is small compared with that spread. \hfill$\square$

\subsection{Proof of \texorpdfstring{\cref{prop:nullspace}}{Proposition \getrefnumber{prop:nullspace}}}
The auditor's output is $p_\theta(\cdot\mid f(m),q)$, so $\Gamma(m,c)$ depends on $m$ only through $f(m)$. For $\delta\in\ker f$, linearity gives $f(m+\delta)=f(m)+f(\delta)=f(m)$, hence $\Gamma(m+\delta,c)=\Gamma(m,c)$ for every question draw. The map in \cref{eq:feat} sends $m$ to $\mathbb{R}^{512|\mathcal{L}|R}$, so its rank is at most $512|\mathcal{L}|R$ and by rank--nullity $\dim\ker f\ge\dim m-512|\mathcal{L}|R$. Layers outside $\mathcal{L}$ do not enter $f$ at all, so perturbations supported on them also lie in $\ker f$. A secret random projection $P$ does not remove the null space: for each draw of $P$ the map is still linear with the same rank bound, and a perturbation that is zero after segment pooling, for instance one that sums to zero over the tokens of every segment, lies in $\ker f$ for every $P$. Our null-space perturbations in \cref{sec:robust} are of this pooling-invariant kind. Reading the receiver's state replaces $f$ by a nonlinear function of $m$, to which the proposition does not apply, and this matches the drop in failure rate against null-space perturbations from 69\% to 58\%. \hfill$\square$

\begingroup\fontsize{8.5}{9.8}\selectfont
\section{Source protocols}

\paragraph{Reader names.} Tables use the reader names of the main paper: question only, linear probe, MLP probe (\emph{prefix}: over the whole latent prefix), attention probe, LTO (the LTO-style reward model), verbalizer (the LatentQA-style 1.7B verbalizer) and \clr{}. Two \clr{} variants appear in the search and design tables: \clr{}+enc.\ uses a pretrained text encoder, and \clr{}+rank is fine-tuned with an additional sibling-ranking loss. Question types follow \cref{sec:labels}: Q1 value, Q2 early exit, Q3 final answer and Q4 intermediate value. Audit sub-questions follow \cref{sec:audit-method}: A1 fact, A2 support, A3 consistency and A4 private value.

\par\vspace{3pt}\begingroup\centering\fontsize{7.5}{8.4}\selectfont\setlength{\tabcolsep}{2.8pt}\renewcommand{\arraystretch}{1}
{\centering\noindent\textbf{Source replay on the original test panel}\par}\nobreak\vspace{1pt}
\noindent\begin{tabular}{@{}lrrr@{}}\toprule
Source / dtype & Batch & Accuracy \% & Agreement \% \\\midrule
GPT-2 bf16 & 128 & 43.594 [40.864,46.247] & 99.924 [99.773,100] \\
GPT-2 bf16 & 1 & 43.518 [40.864,46.247] & 95.603 [94.466,96.664] \\
GPT-2 fp32 & 1 & 43.745 [41.092,46.399] & 95.299 [94.162,96.437] \\
LLaMA bf16 & 16 & 55.269 [52.616,57.923] & 97.953 [97.193,98.711] \\
LLaMA bf16 & 128 & 55.421 [52.767,58.074] & 98.939 [98.332,99.469] \\
\bottomrule\end{tabular}
\par\endgroup\vspace{3pt}
The panel has 1319 problems, and intervals are 95\% bootstrap intervals over problems with 10000 resamples. At batch 1, GPT-2 fp32 and bf16 agree on 1249 problems (94.693\%). The single remaining bf16 discrepancy traces to the first projection path. All experiments use fp32 at batch 1.\par
The official LLaMA accuracy at batch 128 is 55.497\%. Raising our interface batch from 16 to 128 cuts the disagreements from 27 to 14, so about half of the residual disagreement is batch-dependent numerics. We use the released checkpoint as published.\par
\par\vspace{3pt}\begingroup\centering\fontsize{8}{9}\selectfont\setlength{\tabcolsep}{2.8pt}\renewcommand{\arraystretch}{1}
{\centering\noindent\textbf{Train versus held-out overlap}\par}\nobreak\vspace{1pt}
\noindent\begin{tabular}{@{}lrrrr@{}}\toprule
Split & Original & Exact & 13-gram & Removed \\\midrule
val & 250 & 0 & 3 & 3 \\
cal & 250 & 0 & 4 & 4 \\
test & 1319 & 0 & 7 & 7 \\
\bottomrule\end{tabular}
\par\endgroup\vspace{3pt}
The filter normalizes text (NFKC, case folding, quotes and punctuation, keeping signed decimals) and removes every problem with an exact or contiguous 13-token match in the released training corpus. The retained pool is val 300, cal 193 and test 1312. The filter is conservative: it removes every problem with textual overlap, whether or not that problem was in the checkpoint's training data.\par
\section{Noise selection and value labels}

\par\vspace{3pt}\begingroup\centering\fontsize{8}{9}\selectfont\setlength{\tabcolsep}{2.8pt}\renewcommand{\arraystretch}{1}
{\centering\noindent\textbf{GPT-2 noise scan: accuracy / nondegenerate percentage}\par}\nobreak\vspace{1pt}
\noindent\begin{tabular}{@{}lrrrr@{}}\toprule
$\sigma$ & Start 0 acc. & ND & Start 3 acc. & ND \\\midrule
0 & 42.33 & 0.00 & 42.33 & 0.00 \\
0.1 & 42.44 & 1.44 & 42.39 & 1.22 \\
0.3 & 42.45 & 6.67 & 42.47 & 5.33 \\
0.5 & 41.88 & 11.22 & 42.33 & 8.67 \\
0.75 & 41.17 & 19.22 & 42.00 & 14.33 \\
1 & 39.24 & 28.44 & 41.17 & 20.00 \\
1.5 & 30.04 & 44.89 & 37.39 & 32.89 \\
2 & 18.13 & 42.44 & 32.10 & 38.67 \\
Temp. 0.5 & 42.23 & 17.11 & NA & NA \\
Temp. 1 & 40.10 & 31.78 & NA & NA \\
\bottomrule\end{tabular}
\par\endgroup\vspace{3pt}
\par\vspace{3pt}\begingroup\centering\fontsize{8}{9}\selectfont\setlength{\tabcolsep}{2.8pt}\renewcommand{\arraystretch}{1}
{\centering\noindent\textbf{LLaMA noise scan, all relative steps}\par}\nobreak\vspace{1pt}
\noindent\begin{tabular}{@{}lrrr@{}}\toprule
$\sigma$ & Accuracy \% & Drop pp & ND \% \\\midrule
0.5 & 53.88 & 0.45 & 3.06 \\
1.0 & 53.58 & 0.75 & 6.28 \\
1.5 & 53.13 & 1.20 & 10.78 \\
2.0 & 52.34 & 1.99 & 14.78 \\
3.0 & 49.02 & 5.31 & 24.72 \\
4.0 & 43.83 & 10.50 & 35.17 \\
6.0 & 36.02 & 18.31 & 41.11 \\
8.0 & 32.60 & 21.73 & 41.50 \\
\bottomrule\end{tabular}
\par\endgroup\vspace{3pt}
The preregistered rule picks the largest nondegenerate fraction among settings that lose less than 5 pp of accuracy and are at least 30\% nondegenerate, breaking ties toward the smaller $\sigma$. When no setting meets both conditions, the rule keeps the accuracy constraint and takes the largest nondegenerate fraction under it. For LLaMA, whose latent trajectories are more deterministic, this gives $\sigma=2$ with 14.8\% nondegenerate values. A label is nondegenerate when $0<K\widehat V< K$, and the fraction weights relative steps 0--5 equally. The LLaMA scan uses val 300, $K=16$ and seed 11.\par
The GPT-2 scan uses val 300, $K=16$ and seeds 11/22/33. The selected setting, $\sigma=1.5$ from start 3, loses 4.944 pp, which is 0.056 pp below the limit on the seed-mean drop. Intervals for stochastic settings describe variation across seed means; intervals for deterministic settings resample problems. The temperature rows are controls and are not candidates under the registered rule.\par
\par\vspace{3pt}\begingroup\centering\fontsize{7.5}{8.4}\selectfont\setlength{\tabcolsep}{2.8pt}\renewcommand{\arraystretch}{1}
{\centering\noindent\textbf{Value-label histogram by step}\par}\nobreak\vspace{1pt}
\noindent\begin{tabular}{@{}lrrrr@{}}\toprule
Split & Policy & $t$ & $\overline V$ & Zero / interior / one \% \\\midrule
val & $\pi$ & 0 & 0.295 & 48.7/43.0/8.3 \\
 &  & 1 & 0.329 & 47.0/41.7/11.3 \\
 &  & 2 & 0.354 & 47.3/35.0/17.7 \\
 &  & 3 & 0.375 & 47.3/30.7/22.0 \\
 &  & 4 & 0.400 & 49.3/22.0/28.7 \\
 &  & 5 & 0.420 & 55.3/8.0/36.7 \\
val & $\pi_T$ & 0 & 0.408 & 43.0/28.7/28.3 \\
 &  & 1 & 0.403 & 41.0/32.0/27.0 \\
 &  & 2 & 0.403 & 43.0/30.7/26.3 \\
 &  & 3 & 0.398 & 43.3/30.3/26.3 \\
 &  & 4 & 0.406 & 42.0/30.0/28.0 \\
 &  & 5 & 0.406 & 40.3/32.3/27.3 \\
cal & $\pi$ & 0 & 0.304 & 46.0/42.8/11.2 \\
 &  & 1 & 0.336 & 47.2/40.0/12.8 \\
 &  & 2 & 0.365 & 44.8/37.2/18.0 \\
 &  & 3 & 0.388 & 45.4/33.8/20.8 \\
 &  & 4 & 0.412 & 47.0/24.6/28.4 \\
 &  & 5 & 0.445 & 52.2/8.6/39.2 \\
cal & $\pi_T$ & 0 & 0.411 & 42.0/29.8/28.2 \\
 &  & 1 & 0.416 & 40.2/30.4/29.4 \\
 &  & 2 & 0.416 & 39.0/32.0/29.0 \\
 &  & 3 & 0.413 & 40.0/32.0/28.0 \\
 &  & 4 & 0.418 & 39.4/30.6/30.0 \\
 &  & 5 & 0.415 & 38.6/32.2/29.2 \\
test & $\pi$ & 0 & 0.287 & 48.4/41.3/10.3 \\
 &  & 1 & 0.315 & 48.2/37.6/14.2 \\
 &  & 2 & 0.337 & 47.9/35.5/16.6 \\
 &  & 3 & 0.358 & 48.8/30.2/21.0 \\
 &  & 4 & 0.390 & 50.3/22.8/26.9 \\
 &  & 5 & 0.412 & 55.1/10.1/34.7 \\
test & $\pi_T$ & 0 & 0.387 & 41.1/31.6/27.3 \\
 &  & 1 & 0.384 & 43.5/29.3/27.3 \\
 &  & 2 & 0.387 & 43.1/29.9/27.1 \\
 &  & 3 & 0.384 & 43.5/29.6/27.0 \\
 &  & 4 & 0.387 & 42.4/31.0/26.6 \\
 &  & 5 & 0.386 & 41.9/31.4/26.7 \\
\bottomrule\end{tabular}
\par\endgroup\vspace{3pt}
Labels take values in $\{0,1/16,\ldots,1\}$. The table merges the 15 interior bins and reports the two endpoint masses separately. Each step has 300, 500 or 1005 problems for val, cal and test. $\pi_T$ also samples the answer decoder, so its continuation target differs from that of $\pi$.\par
\par\vspace{3pt}\begingroup\centering\fontsize{8}{9}\selectfont\setlength{\tabcolsep}{2.8pt}\renewcommand{\arraystretch}{1}
{\centering\noindent\textbf{Variance of a K16 label, estimated from K64}\par}\nobreak\vspace{1pt}
\noindent\begin{tabular}{@{}lrrr@{}}\toprule
Policy & Pairs & Mean variance & 95\% interval \\\midrule
$\pi$ & 50 & 0.002585 & [0.001139,0.004382] \\
$\pi_T$ & 50 & 0.003424 & [0.001895,0.005293] \\
\bottomrule\end{tabular}
\par\endgroup\vspace{3pt}
Each of the 50 problem-step pairs contributes four independent K16 blocks. They use seeds different from those of the primary labels. The quantity estimated is the sampling variance of one K16 label; the variance of a K64 mean is smaller.\par
\clearpage
\section{Question bank}

\par\vspace{3pt}\begingroup\fontsize{8}{9}\selectfont\setlength{\tabcolsep}{2.8pt}\renewcommand{\arraystretch}{1}
{\centering\noindent\textbf{Question templates (examples)}\par}\nobreak\vspace{1pt}
\noindent\begin{tabularx}{\linewidth}{@{}llX@{}}\toprule
Type & Template & Question \\\midrule
Q1 & Seen 0 & What is the probability that continuing from this state ends with a correct answer? \\
Q1 & Seen 1 & Estimate the chance of a correct final answer from this state. \\
Q1 & Held-out 4 & If reasoning continues from here, what is its chance of success? \\
Q2 & Seen 0 & Would an answer forced at this state be correct? \\
Q2 & Seen 1 & If reasoning stops now, will the answer be correct? \\
Q2 & Held-out 4 & Does an early exit here produce a correct answer? \\
Q3 & Seen 0 & Which option is the final answer the model will produce? \\
Q3 & Seen 1 & Select the answer that this reasoning will end with. \\
Q3 & Held-out 4 & Choose the option matching the eventual output. \\
Q4 & Seen 0 & Is $v$ a quantity appearing in the standard equation solution? \\
Q4 & Seen 1 & Does the standard solution contain the computed value $v$? \\
Q4 & Held-out 4 & Does a reference equation yield the value $v$? \\
\bottomrule\end{tabularx}
\par\endgroup\vspace{3pt}
Templates 0--3 are used for training and templates 4--5 are held out. Each evaluation instance is asked with one seen paraphrase and with both held-out ones. The reader sees the observed state, the question text and the options, and never the source problem or the gold reasoning.\par
\par\vspace{3pt}\begingroup\fontsize{8}{9}\selectfont\setlength{\tabcolsep}{2.8pt}\renewcommand{\arraystretch}{1}
{\centering\noindent\textbf{Answer construction and supervision}\par}\nobreak\vspace{1pt}
\noindent\begin{tabularx}{\linewidth}{@{}lX@{}}\toprule
Type & Construction \\\midrule
Q1 & Bernoulli continuation outcomes; train labels begin at the trajectory start. Earlier clean prefixes receive no label. Cal/test retain each K16 outcome. \\
Q2 & Binary correctness of clean immediate answer decoding. \\
Q3 & Four normalized numeric options, containing the actual final model answer. Distractors use same-problem behavior, gold and a fixed training-answer pool. Invalid answers remain explicit. \\
Q4 & Membership in equation right-hand-side values, including the final calculation. Within-split derangement preserves positive and negative numeric marginals and excludes the true value. \\
\bottomrule\end{tabularx}
\par\endgroup\vspace{3pt}
\par\vspace{3pt}\begingroup\centering\fontsize{7.5}{8.4}\selectfont\setlength{\tabcolsep}{2.2pt}\renewcommand{\arraystretch}{1}
{\centering\noindent\textbf{Dataset scale and answer-position frequencies}\par}\nobreak\vspace{1pt}
\noindent\begin{tabular}{@{}lrrrrr@{}}\toprule
Split & Problems & QA rows & Parse & Unpaired & Q3 positions 0 / 1 / 2 / 3 \% \\\midrule
reader-train & 20000 & 2019896 & 40 & 0 & 24.96/24.92/25.03/25.09 \\
reader-val & 2000 & 606204 & 3 & 1 & 25.17/24.82/25.05/24.96 \\
val & 300 & 31332 & 3 & 1 & 24.05/24.67/23.76/27.52 \\
cal & 500 & 52080 & 9 & 1 & 24.60/24.89/25.63/24.89 \\
test & 1005 & 104979 & 12 & 1 & 24.63/25.13/24.95/25.29 \\
\bottomrule\end{tabular}
\par\endgroup\vspace{3pt}
The Q4 positive rate is exactly 50\% in each original split. The two cal halves are assigned by an independent hash and have 250 problems each, so their value marginals can deviate slightly. Option order follows a fixed random permutation, which is why a finite panel does not place exactly 25\% of answers at each position.\par
\par\vspace{3pt}\begingroup\centering\fontsize{8}{9}\selectfont\setlength{\tabcolsep}{2.8pt}\renewcommand{\arraystretch}{1}
{\centering\noindent\textbf{GPT-2 clean Q1 soft mean / Q2 positive rate in percent}\par}\nobreak\vspace{1pt}
\noindent\begin{tabular}{@{}lrrrr@{}}\toprule
$t$ & Train & Val & Cal & Test \\\midrule
0 & 67.0/49.3 & 29.5/19.7 & 30.4/25.4 & 28.7/23.3 \\
1 & 69.8/59.0 & 32.9/28.0 & 33.6/24.8 & 31.5/24.7 \\
2 & 73.6/58.0 & 35.4/24.0 & 36.5/23.2 & 33.7/22.7 \\
3 & 76.3/77.4 & 37.5/39.3 & 38.8/37.8 & 35.8/36.4 \\
4 & 78.8/76.7 & 40.0/38.3 & 41.2/37.8 & 39.0/36.7 \\
5 & 81.2/82.0 & 42.0/43.0 & 44.5/45.0 & 41.2/41.6 \\
6 & 81.9/81.9 & 42.3/42.3 & 45.2/45.2 & 42.0/42.0 \\
\bottomrule\end{tabular}
\par\endgroup\vspace{3pt}
\section{Reader implementation and diagnostics}

\par\vspace{3pt}\begingroup\fontsize{8}{9}\selectfont\setlength{\tabcolsep}{2.8pt}\renewcommand{\arraystretch}{1}
{\centering\noindent\textbf{Architectures and optimization}\par}\nobreak\vspace{1pt}
\noindent\begin{tabularx}{\linewidth}{@{}lX@{}}\toprule
Reader & Specification \\\midrule
\clr{} & 4 layers, width 256, 8 heads, FFN width 1024, zero dropout; ordered state prefix, typed text, numeric features; joint width-256 head. \\
\clr{}+enc. & ModernBERT-base, all 22 blocks; cross-attention every 4 blocks; initial residual scale 0.1; joint option/state/numeric head. \\
Shared training & AdamW, new-module LR $3\times10^{-4}$, encoder LR $10^{-5}$ for \clr{}+enc., WD $10^{-4}$; batch 1024 for \clr{} and 256 for \clr{}+enc. Full traversal with inverse-type weights; checkpoint epochs 8--10, patience 1. \\
MLP probe, prefix & Independent per type, hidden width 128, batch 4096, 8 epochs, LR grid $10^{-3},3\times10^{-4}$, WD $10^{-4}$. Prefix concatenates normalized prefix mean and current state. \\
Attention probe & One cross-attention and linear head per type; width 256, 8 heads, batch 4096, 8 epochs; LR grid $10^{-3},3\times10^{-4}$, WD $10^{-4}$; frozen option embeddings. \\
Verbalizer & Frozen Qwen3-1.7B, two-layer latent injection, rank-16 LoRA with alpha 32; batch 512, LR $2\times10^{-4}$, WD 0.01; one complete pass over GPT-2 Q3/Q4, all three seeds. \\
\bottomrule\end{tabularx}
\par\endgroup\vspace{3pt}
All main reader comparisons use seeds 11/22/33. The verbalizer is trained on 860228 Q3 and 279440 Q4 rows per seed and evaluated on GPT-2 Q3/Q4, the questions on which the verbalizer comparison is made. Because it differs from \clr{} in both training exposure and architecture, the comparison is between complete systems.\par
\par\vspace{3pt}\begingroup\fontsize{8}{9}\selectfont\setlength{\tabcolsep}{2.8pt}\renewcommand{\arraystretch}{1}
{\centering\noindent\textbf{Reader design diagnostics}\par}\nobreak\vspace{1pt}
\noindent\begin{tabularx}{\linewidth}{@{}lX@{}}\toprule
Check & Finding \\\midrule
Pathways & Nonzero state gradients and centered option-logit permutation effects for every type. Measured in a diagnostic replay. \\
Masks & Each cross-attention layer exposes valid states; padding is excluded and options are isolated. The initial reader has no gate. \\
Exposure & The initial reader selected 40000 / 40000 / 60000 examples per type across seeds; MLP probe Q3: 4301140 / 5161368 / 5161368; Q4: 2235520 each. Exposure ratios span about 37--129. \\
Memorization & On the same 1000-example Q4 diagnostic set: MLP probe 98.0\%, initial reader 81.2\%, initial reader continued at lower LR 97.0\%. Both final candidates pass the 98\% gate. \\
Synthetic & Unit-L2 linear integer encoding: MLP probe Q3/Q4 98.1/98.2\%; initial reader 98.6/99.6\%. Both final candidates pass each 98\% validation gate. \\
\bottomrule\end{tabularx}
\par\endgroup\vspace{3pt}
The final configuration adds a cross residual with a nonzero, controlled scale and an explicit nonlinear head over state and option, lowers the learning rate of the new modules, and removes the per-type sampling cap, so that \clr{} receives at least the training exposure of the probes. These changes act jointly; the checks verify each pathway rather than isolate one change.\par
\clearpage
\noindent\textbf{Reader results, continued.} All scores below are computed from the frozen test predictions.\par
\par\vspace{3pt}\begingroup\centering\fontsize{7.5}{8.4}\selectfont\setlength{\tabcolsep}{2.8pt}\renewcommand{\arraystretch}{1}
{\centering\noindent\textbf{Training exposure and updates of the selected checkpoints}\par}\nobreak\vspace{1pt}
\noindent\begin{tabular}{@{}lrrrrrr@{}}\toprule
Model & Seed & Q1 value & Q2 exit & Q3 answer & Q4 interm. & Updates \\\midrule
\clr{} & 11 & 5921824 & 1120000 & 6881824 & 2235520 & 15792 \\
\clr{} & 22 & 5921824 & 1120000 & 6881824 & 2235520 & 15792 \\
\clr{} & 33 & 5921824 & 1120000 & 6881824 & 2235520 & 15792 \\
\clr{}+enc. & 11 & 5921824 & 1120000 & 6881824 & 2235520 & 63136 \\
\clr{}+enc. & 22 & 5921824 & 1120000 & 6881824 & 2235520 & 63136 \\
\clr{}+enc. & 33 & 5921824 & 1120000 & 6881824 & 2235520 & 63136 \\
\bottomrule\end{tabular}
\par\endgroup\vspace{3pt}
\par\vspace{3pt}\begingroup\centering\fontsize{8}{9}\selectfont\setlength{\tabcolsep}{2.8pt}\renewcommand{\arraystretch}{1}
{\centering\noindent\textbf{Shared versus independent per-type \clr{}}\par}\nobreak\vspace{1pt}
\noindent\begin{tabular}{@{}lrrr@{}}\toprule
Source & Type & Multi-type & Per-type \\\midrule
GPT-2 & Q1 value & 0.558 [0.535,0.581] & 0.557 [0.536,0.579] \\
 & Q2 exit & 0.831 [0.812,0.849] & 0.823 [0.804,0.842] \\
 & Q3 answer & 76.2 [74.9,77.4] & 76.7 [75.4,77.9] \\
 & Q4 interm. & 76.4 [75.0,77.9] & 73.7 [72.1,75.3] \\
LLaMA & Q1 value & 0.622 [0.603,0.640] & 0.634 [0.616,0.651] \\
 & Q2 exit & 0.798 [0.778,0.817] & 0.818 [0.799,0.837] \\
 & Q3 answer & 75.0 [73.5,76.4] & 72.6 [71.0,74.1] \\
 & Q4 interm. & 76.5 [75.0,77.9] & 71.4 [69.7,73.0] \\
\bottomrule\end{tabular}
\par\endgroup\vspace{3pt}
Each per-type model has the full \clr{} architecture and sees exactly as many examples of its type as the shared model saw in 8 complete epochs. Intervals hold the three training seeds fixed and resample whole problems.\par
\par\vspace{3pt}\begingroup\centering\fontsize{8}{9}\selectfont\setlength{\tabcolsep}{2.8pt}\renewcommand{\arraystretch}{1}
{\centering\noindent\textbf{Ablations on CODI-LLaMA}\par}\nobreak\vspace{1pt}
\noindent\begin{tabular}{@{}lrrrrr@{}}\toprule
Variant & Q1 NLL$\downarrow$ & Q2 AUC$\uparrow$ & Q3$\uparrow$ & Q4$\uparrow$ & $|C_{\mathrm{Q3}}|\downarrow$ \\\midrule
\clr{} & 0.615 & 0.821 & 72.1 & 74.2 & 1.64 \\
w/o block mask & 0.619$^\dagger$ & 0.811$^\dagger$ & 64.0$^\dagger$ & 71.9$^\dagger$ & 1.91$^\dagger$ \\
dot-product head & 0.627$^\dagger$ & 0.786$^\dagger$ & 68.4$^\dagger$ & 73.3 & 1.78$^\dagger$ \\
late fusion & 0.628$^\dagger$ & 0.819 & 75.7$^\ddagger$ & 76.4$^\ddagger$ & 1.45$^\ddagger$ \\
single step $z_t$ & 0.627$^\dagger$ & 0.813 & 74.7$^\ddagger$ & 75.5$^\ddagger$ & 1.50$^\ddagger$ \\
one model per type & 0.634$^\dagger$ & 0.818 & 72.6 & 71.4$^\dagger$ & 1.64 \\
\bottomrule\end{tabular}
\par\endgroup\vspace{3pt}
Same protocol as the GPT-2 ablations: identical training recipe and budget with only the source width changed, three seeds, all variants frozen before the test set is opened, and the same calibration. $^\dagger$/$^\ddagger$: worse/better than \clr{} with a paired 95\% interval that excludes zero; the per-type models use matched exposure. The block mask and the joint state--option head help on both sources, and on LLaMA by more (8.1 and 3.7 Q3 points), so the option--token interaction is the part of the design that carries over. Multi-type training improves Q4 by 2.8 points on LLaMA and 2.5 on GPT-2, with no detectable change on Q3 on either. Fusion depth and prefix length behave as per-source settings: late fusion and the single-step input are stronger on LLaMA and weaker on GPT-2. The main tables use one fixed configuration for both sources; checkpoints and the choice between the from-scratch reader and \clr{}+enc.\ were made on reader-validation NLL, and the single-step and source-joint rows of \cref{tab:main} were evaluated after this configuration was fixed.\par
\par\vspace{3pt}\begingroup\centering\fontsize{8}{9}\selectfont\setlength{\tabcolsep}{2.8pt}\renewcommand{\arraystretch}{1}
{\centering\noindent\textbf{\clr{} against the LatentQA-style verbalizer on GPT-2}\par}\nobreak\vspace{1pt}
\noindent\begin{tabular}{@{}llrrc@{}}\toprule
Type & Metric & \clr{}$-$verb., 95\% & 90\% interval & Margin \\\midrule
Q3 & accuracy (pt) & $-$0.66 [$-$1.51, 0.19] & [$-$1.38, 0.05] & $\pm$2.0 \\
Q4 & accuracy (pt) & $-$0.09 [$-$0.98, 0.83] & [$-$0.85, 0.67] & $\pm$2.0 \\
Q3 & set size & +0.017 [0.007, 0.027] & [0.008, 0.025] & $\pm$0.10 \\
Q4 & set size & +0.019 [0.006, 0.033] & [0.009, 0.031] & $\pm$0.10 \\
\bottomrule\end{tabular}
\par\endgroup\vspace{3pt}
Equivalence is tested with two one-sided tests at level 0.05: the paired 90\% bootstrap interval must lie inside the margin, which was fixed before the test was run. All four comparisons pass. The accuracy intervals would pass margins of $\pm$1.4 (Q3) and $\pm$0.9 (Q4) points. \clr{}'s sets are larger by less than 0.02 options, a difference that is statistically detectable and well inside the margin.\par
\par\vspace{3pt}\begingroup\centering\fontsize{7.5}{8.4}\selectfont\setlength{\tabcolsep}{1.3pt}\renewcommand{\arraystretch}{1}
{\centering\noindent\textbf{\clr{} against single-type readers on state-only questions}\par}\nobreak\vspace{1pt}
\noindent\begin{tabular}{@{}llrr@{}}\toprule
Source & \clr{} $-$ & $\Delta$Q1 NLL & $\Delta$Q2 AUROC \\\midrule
GPT-2 & MLP probe & +0.026 [0.015, 0.037] & $-$0.017 [$-$0.026, $-$0.007] \\
 & \shortstack[l]{MLP probe,\\prefix} & +0.033 [0.022, 0.044] & $-$0.028 [$-$0.038, $-$0.018] \\
 & \shortstack[l]{Attention\\probe} & $-$0.026 [$-$0.038, $-$0.013] & $-$0.023 [$-$0.033, $-$0.014] \\
 & LTO & +0.042 [0.029, 0.054] & $-$0.015 [$-$0.027, $-$0.004] \\
 & HSRM-style & $-$0.011 [$-$0.022, 0.001] & $-$0.005 [$-$0.016, 0.006] \\
 & CLUE & $-$0.100 [$-$0.129, $-$0.072] & +0.254 [0.226, 0.282] \\
\midrule
LLaMA & MLP probe & +0.004 [$-$0.002, 0.009] & $-$0.030 [$-$0.040, $-$0.020] \\
 & \shortstack[l]{MLP probe,\\prefix} & +0.009 [0.003, 0.015] & $-$0.031 [$-$0.041, $-$0.021] \\
 & \shortstack[l]{Attention\\probe} & $-$0.017 [$-$0.024, $-$0.009] & $-$0.002 [$-$0.014, 0.009] \\
 & LTO & +0.064 [0.035, 0.091] & +0.016 [$-$0.0003, 0.031] \\
 & HSRM-style & $-$0.009 [$-$0.015, $-$0.004] & $-$0.013 [$-$0.026, $-$0.0001] \\
 & CLUE & +0.071 [0.051, 0.090] & +0.029 [0.014, 0.043] \\
\bottomrule\end{tabular}
\par\endgroup\vspace{3pt}
Paired differences with 95\% intervals; lower NLL and higher AUROC are better. Each of these readers is trained or built for one or two state-only questions, while \clr{} answers all four types with one model. \clr{} is ahead of the attention probe on Q1 on both sources and of CLUE on GPT-2, and ahead of the HSRM-style reader on LLaMA Q1. The best single-type reader in each column leads by at most 0.071 NLL (CLUE, LLaMA) and 0.031 AUROC (MLP probe with prefix, LLaMA). CLUE is deterministic and is paired with \clr{} on the same resampled problems.\par
\par\vspace{3pt}\begingroup\centering\fontsize{8}{9}\selectfont\setlength{\tabcolsep}{2.8pt}\renewcommand{\arraystretch}{1}
{\centering\noindent\textbf{GPT-2 clean seen-template results at every step}\par}\nobreak\vspace{1pt}
\noindent\begin{tabular}{@{}lrrrrr@{}}\toprule
Reader & $t$ & Q1 value & Q2 exit & Q3 answer & Q4 interm. \\
 & & NLL & AUROC & \% & \% \\\midrule
MLP probe & 0 & 0.599 & 0.711 & 49.8 & 67.2 \\
 & 1 & 0.492 & 0.833 & 60.8 & 69.0 \\
 & 2 & 0.553 & 0.861 & 53.4 & 67.7 \\
 & 3 & 0.495 & 0.901 & 66.5 & 69.5 \\
 & 4 & 0.516 & 0.877 & 56.0 & 67.9 \\
 & 5 & 0.498 & 0.884 & 68.4 & 68.7 \\
 & 6 & 0.506 & 0.856 & 55.3 & 67.7 \\
\clr{} & 0 & 0.607 & 0.726 & 63.0 & 72.2 \\
 & 1 & 0.546 & 0.803 & 74.3 & 74.2 \\
 & 2 & 0.545 & 0.848 & 75.7 & 76.8 \\
 & 3 & 0.530 & 0.869 & 79.3 & 77.3 \\
 & 4 & 0.528 & 0.866 & 79.2 & 77.7 \\
 & 5 & 0.538 & 0.848 & 80.8 & 77.4 \\
 & 6 & 0.542 & 0.847 & 80.4 & 77.7 \\
\clr{}+enc. & 0 & 0.606 & 0.726 & 64.6 & 71.8 \\
 & 1 & 0.561 & 0.802 & 74.2 & 74.1 \\
 & 2 & 0.560 & 0.848 & 75.3 & 75.2 \\
 & 3 & 0.555 & 0.864 & 79.6 & 75.6 \\
 & 4 & 0.547 & 0.866 & 79.8 & 76.4 \\
 & 5 & 0.561 & 0.851 & 79.2 & 75.8 \\
 & 6 & 0.559 & 0.855 & 80.9 & 76.3 \\
\bottomrule\end{tabular}
\par\endgroup\vspace{3pt}
\par\vspace{3pt}\begingroup\centering\fontsize{8}{9}\selectfont\setlength{\tabcolsep}{2.8pt}\renewcommand{\arraystretch}{1}
{\centering\noindent\textbf{GPT-2 template generalization, equal weight over steps}\par}\nobreak\vspace{1pt}
\noindent\begin{tabular}{@{}lrrrrr@{}}\toprule
Reader & Template & Q1 value & Q2 exit & Q3 answer & Q4 interm. \\
 & & NLL & AUROC & \% & \% \\\midrule
MLP probe & Seen & 0.523 & 0.846 & 58.6 & 68.3 \\
MLP probe & Held-out & 0.522 & 0.846 & 58.7 & 68.3 \\
\clr{} & Seen & 0.548 & 0.830 & 76.1 & 76.2 \\
\clr{} & Held-out & 0.552 & 0.829 & 72.4 & 74.5 \\
\clr{}+enc. & Seen & 0.564 & 0.830 & 76.2 & 75.0 \\
\clr{}+enc. & Held-out & 0.565 & 0.830 & 75.8 & 74.9 \\
\bottomrule\end{tabular}
\par\endgroup\vspace{3pt}
Temperatures are fitted on $\mathrm{cal}_A$, and the held-out rows reuse the calibrator fitted on seen templates. Coverage on held-out templates and under shift is measured empirically, since these settings lie outside the exchangeability assumption. Training passes over all instances, which keeps their natural imbalance across types, and the loss compensates with inverse-type weights.\par
\par\vspace{3pt}\begingroup\centering\fontsize{8}{9}\selectfont\setlength{\tabcolsep}{2.8pt}\renewcommand{\arraystretch}{1}
{\centering\noindent\textbf{Pathway diagnostics across all cases and question types}\par}\nobreak\vspace{1pt}
\noindent\begin{tabular}{@{}lrr@{}}\toprule
Diagnostic & Minimum & Maximum \\\midrule
State gradient norm & 0.01512 & 98.812 \\
Centered gradient norm & 0.0067 & 49.406 \\
Shuffle centered logits & 0.0056528 & 3.0861 \\
\bottomrule\end{tabular}
\par\endgroup\vspace{3pt}
The bf16 replay also shows nonzero state gradients and permutation effects in all 80 batches, and the MLP probe and \clr{} share the final encoding control. The final configuration was designed after test results of the initial reader had been computed; checkpoints are selected on reader-validation NLL only.\par
\par\vspace{3pt}\begingroup\fontsize{8}{9}\selectfont\setlength{\tabcolsep}{2.8pt}\renewcommand{\arraystretch}{1}
{\centering\noindent\textbf{Runtime and FLOP measurement protocol}\par}\nobreak\vspace{1pt}
\noindent\begin{tabularx}{\linewidth}{@{}lX@{}}\toprule
Item & Convention \\\midrule
Timing & Seed-11 weights, shared RTX 4090, 64 reader-val problems; median of 3 timing rounds, 2 warmups and 3 inner loops. Feature preparation, tokenization and transfers are excluded. \\
FLOPs & Matrix and attention operations use 2 FLOPs/MAC; lookups, normalization, nonlinearities and softmax are excluded. Batch padding changes cost. \\
Storage & The attention probe includes 38.7M frozen embedding parameters; the verbalizer includes its 1.74B frozen decoder. The reported training costs cover Q1--Q4 for \clr{} and the attention probe and Q3--Q4 once for the verbalizer; costs repeat across question types. \\
\bottomrule\end{tabularx}
\par\endgroup\vspace{3pt}
\par\vspace{3pt}\begingroup\centering\fontsize{8}{9}\selectfont\setlength{\tabcolsep}{2.8pt}\renewcommand{\arraystretch}{1}
{\centering\noindent\textbf{Parameter counts of the profiled readers}\par}\nobreak\vspace{1pt}
\noindent\begin{tabular}{@{}lrr@{}}\toprule
Reader & Trainable & Total \\\midrule
MLP probe & 107777 & 107777 \\
Attention probe & 661249 & 39343873 \\
\clr{} & 17180931 & 17180931 \\
Verbalizer & 18876928 & 1739451904 \\
\bottomrule\end{tabular}
\par\endgroup\vspace{3pt}
\endgroup\onecolumn\begingroup\fontsize{8.5}{9.8}\selectfont
\section{Search settings and within-group diagnostics}
\begin{figure*}[t]
\centering
\includegraphics[width=\textwidth]{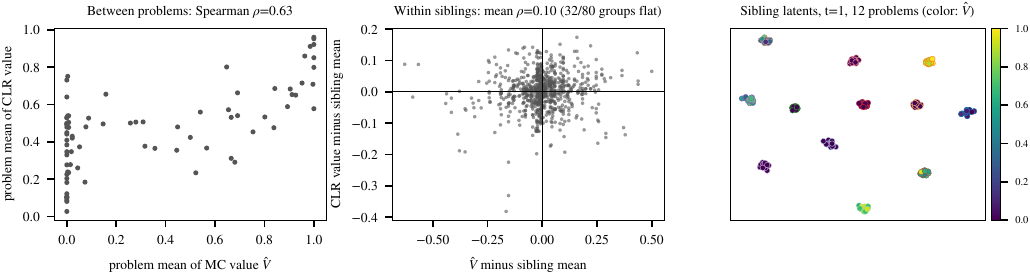}
\caption{Between-problem and within-sibling structure of \clr{}'s Q1 value on 40 validation problems at $t\in\{1,3\}$, 17 children per parent including the noise-free child. Left: problem means of \clr{}'s value against problem means of the Monte Carlo value. Middle: both quantities centered within each sibling group; groups in which all children have the same Monte Carlo value are omitted. Right: t-SNE of the sibling latents for 12 problems at $t=1$, colored by Monte Carlo value. Siblings form tight clusters, one per problem, and their values vary inside a cluster in ways the reader does not track. This is the situation of \cref{prop:within}.}
\label{fig:qual-siblings}
\end{figure*}

\par\vspace{3pt}\begingroup\centering\fontsize{8}{9}\selectfont\setlength{\tabcolsep}{2.8pt}\renewcommand{\arraystretch}{1}
{\centering\noindent\textbf{All 90 learned-value settings: accuracy percentage / source-equivalent forwards per problem}\par}\nobreak\vspace{1pt}
\noindent\begin{tabular}{@{}lrrrrrrrr@{}}\toprule
Search & $N$ & $b$ & $h$ & LTO & MLP probe & \clr{} & \clr{}+enc. & \clr{}+rank \\\midrule
Best-of-$N$ & 4 & 1 & 4 & 40.78 / 31.2 & 39.78 / 31.2 & 40.56 / 34.0 & 40.67 / 106.5 & 40.56 / 34.0 \\
Best-of-$N$ & 4 & 1 & 6 & 37.89 / 31.2 & 39.11 / 31.2 & 40.89 / 34.0 & 40.78 / 106.5 & 41.11 / 34.0 \\
Best-of-$N$ & 8 & 1 & 4 & 40.67 / 55.2 & 40.44 / 55.2 & 41.22 / 60.8 & 40.56 / 205.8 & 40.89 / 60.8 \\
Best-of-$N$ & 8 & 1 & 6 & 37.78 / 55.2 & 38.22 / 55.2 & 39.78 / 60.8 & 39.00 / 205.9 & 40.89 / 60.8 \\
Best-of-$N$ & 16 & 1 & 4 & 39.89 / 103.2 & 38.11 / 103.2 & 39.67 / 114.4 & 39.11 / 404.5 & 40.11 / 114.3 \\
Best-of-$N$ & 16 & 1 & 6 & 36.22 / 103.2 & 35.56 / 103.3 & 38.78 / 114.4 & 38.89 / 404.5 & 39.56 / 114.4 \\
Beam & 4 & 2 & 4 & 41.11 / 37.2 & 40.67 / 37.2 & 40.00 / 54.2 & 40.89 / 559.2 & 40.22 / 54.2 \\
Beam & 4 & 2 & 6 & 38.89 / 51.2 & 40.33 / 51.3 & 38.78 / 79.1 & 38.67 / 874.0 & 39.00 / 79.1 \\
Beam & 4 & 4 & 4 & 39.11 / 61.2 & 39.22 / 61.2 & 39.33 / 92.9 & 39.44 / 1031.0 & 40.00 / 92.9 \\
Beam & 4 & 4 & 6 & 39.11 / 91.2 & 41.44 / 91.3 & 40.00 / 144.7 & 38.89 / 1662.7 & 40.78 / 144.8 \\
Beam & 8 & 2 & 4 & 38.56 / 65.2 & 39.00 / 65.3 & 38.22 / 99.1 & 37.11 / 1109.2 & 39.22 / 99.1 \\
Beam & 8 & 2 & 6 & 36.44 / 95.2 & 38.33 / 95.3 & 38.11 / 151.0 & 35.78 / 1740.8 & 37.78 / 151.0 \\
Beam & 8 & 4 & 4 & 40.44 / 113.2 & 39.78 / 113.3 & 38.56 / 176.6 & 37.44 / 2052.8 & 40.56 / 176.6 \\
Beam & 8 & 4 & 6 & 37.11 / 175.2 & 37.33 / 175.4 & 37.33 / 282.3 & 35.44 / 3318.0 & 38.00 / 282.3 \\
Beam & 16 & 2 & 4 & 38.22 / 121.2 & 39.11 / 121.3 & 38.00 / 189.1 & 36.67 / 2209.2 & 38.78 / 189.1 \\
Beam & 16 & 2 & 6 & 35.67 / 183.2 & 37.22 / 183.4 & 35.56 / 294.8 & 34.44 / 3474.4 & 36.89 / 294.8 \\
Beam & 16 & 4 & 4 & 36.67 / 217.2 & 38.33 / 217.4 & 38.00 / 343.9 & 36.89 / 4096.4 & 38.22 / 343.9 \\
Beam & 16 & 4 & 6 & 36.89 / 343.2 & 37.89 / 343.5 & 35.00 / 557.4 & 34.44 / 6628.9 & 37.56 / 557.3 \\
\bottomrule\end{tabular}
\par\endgroup\vspace{3pt}
The panel has 300 validation problems and three seeds. $N$ counts one clean candidate per parent. Best-of-$N$ scores complete trajectories and keeps one full clean trajectory among the candidates; beam search keeps the clean child at every expansion but may prune its clean ancestor. With $h=4$ the search branches up to $t\le3$ and then decodes deterministically; with $h=6$ it branches at every step. Source prefill is counted once per search branch and is not weighted by token length. The cost also includes the matrix and attention operations of the reader in source-forward equivalents. It is a source-forward compute axis, distinct from FLOPs or latency.\par
\par\vspace{3pt}\begingroup\centering\fontsize{7.5}{8.4}\selectfont\setlength{\tabcolsep}{2.4pt}\renewcommand{\arraystretch}{1}
{\centering\noindent\textbf{Every step: cross-problem AUROC / within-parent $\rho$ / large-gap pair accuracy / selected-minus-clean value in pp}\par}\nobreak\vspace{1pt}
\noindent\begin{tabular}{@{}lrrrrrr@{}}\toprule
$t$ & LTO & MLP probe & \clr{} & \clr{}+enc. & \clr{}+rank & Valid $\rho$ groups \\\midrule
0 & 0.843 / 0.079 / 0.586 / -2.77 & 0.848 / 0.107 / 0.617 / -1.57 & 0.805 / 0.090 / 0.588 / -2.00 & 0.792 / 0.078 / 0.593 / -2.64 & 0.811 / 0.108 / 0.607 / -1.70 & 555 \\
1 & 0.835 / 0.093 / 0.599 / -1.75 & 0.839 / 0.115 / 0.631 / -1.27 & 0.818 / 0.097 / 0.620 / -1.63 & 0.815 / 0.092 / 0.621 / -1.28 & 0.827 / 0.095 / 0.629 / -0.99 & 541 \\
2 & 0.854 / 0.088 / 0.578 / -2.24 & 0.863 / 0.108 / 0.591 / -2.01 & 0.829 / 0.110 / 0.616 / -1.76 & 0.819 / 0.111 / 0.588 / -2.01 & 0.835 / 0.117 / 0.627 / -1.26 & 509 \\
3 & 0.858 / 0.124 / 0.657 / -0.97 & 0.857 / 0.129 / 0.635 / -1.12 & 0.833 / 0.126 / 0.602 / -1.03 & 0.830 / 0.132 / 0.623 / -1.07 & 0.837 / 0.132 / 0.603 / -1.16 & 475 \\
4 & 0.864 / 0.082 / 0.576 / -2.35 & 0.869 / 0.111 / 0.603 / -1.60 & 0.834 / 0.111 / 0.591 / -1.50 & 0.833 / 0.092 / 0.578 / -1.11 & 0.842 / 0.116 / 0.592 / -0.93 & 320 \\
5 & 0.856 / 0.144 / 0.638 / 0.33 & 0.854 / 0.175 / 0.675 / 0.22 & 0.843 / 0.157 / 0.649 / 0.11 & 0.839 / 0.163 / 0.658 / 0.33 & 0.848 / 0.180 / 0.681 / 0.33 & 80 \\
\bottomrule\end{tabular}
\par\endgroup\vspace{3pt}
Each parent has one clean and 16 perturbed children, each with its own K16 continuation label. Groups whose true values or predictions are constant are left out of the Spearman correlation. Pair accuracy uses pairs with $|\Delta\widehat V|\ge0.25$ and counts ties as 0.5. Group statistics are averaged within each problem and bootstrapped over problems. In 54.1\% of groups all children have the same true value, and the final rollout at $t=5$ is deterministic.\par
\par\vspace{3pt}\begingroup\centering\fontsize{8}{9}\selectfont\setlength{\tabcolsep}{2.8pt}\renewcommand{\arraystretch}{1}
{\centering\noindent\textbf{Pooled sibling diagnostics with 95\% problem intervals}\par}\nobreak\vspace{1pt}
\noindent\begin{tabular}{@{}lrrrr@{}}\toprule
Value & Cross-problem AUROC & Within $\rho$ & Pair accuracy & Selected $\Delta V$ pp \\\midrule
LTO & 0.852 [0.847,0.856] & 0.084 [0.065,0.103] & 0.611 [0.584,0.636] & -1.62 [-2.19,-1.06] \\
MLP probe & 0.855 [0.854,0.855] & 0.103 [0.083,0.124] & 0.622 [0.595,0.649] & -1.22 [-1.76,-0.72] \\
\clr{} & 0.827 [0.817,0.837] & 0.099 [0.080,0.118] & 0.596 [0.566,0.623] & -1.30 [-1.80,-0.82] \\
\clr{}+enc. & 0.821 [0.798,0.844] & 0.101 [0.081,0.121] & 0.600 [0.573,0.628] & -1.30 [-1.80,-0.81] \\
\clr{}+rank & 0.833 [0.823,0.844] & 0.104 [0.081,0.125] & 0.614 [0.585,0.642] & -0.95 [-1.42,-0.49] \\
\bottomrule\end{tabular}
\par\endgroup\vspace{3pt}
\noindent\begin{minipage}[t]{0.48\linewidth}
\par\vspace{3pt}\begingroup\centering\fontsize{7.5}{8.4}\selectfont\setlength{\tabcolsep}{2.8pt}\renewcommand{\arraystretch}{1}
{\centering\noindent\textbf{Compute-matched reference with oracle}\par}\nobreak\vspace{1pt}
\noindent\begin{tabular}{@{}lrrr@{}}\toprule
Beam, $N{=}16$, $b{=}4$, $h{=}6$ & Accuracy \% & Cost & Gain vs.\ voting pp \\\midrule
MLP probe & 36.4 [31.4,41.7] & 343.5 & -0.7 [-3.8,2.2] \\
LTO & 34.9 [30.1,39.9] & 343.2 & -2.2 [-5.2,0.9] \\
oracle & 70.6 [65.6,75.6] & 45588.6 & 32.8 [27.8,38.0] \\
\bottomrule\end{tabular}
\par\endgroup\vspace{3pt}
\end{minipage}\hfill\begin{minipage}[t]{0.49\linewidth}
\par\vspace{3pt}\begingroup\centering\fontsize{7.5}{8.4}\selectfont\setlength{\tabcolsep}{2.8pt}\renewcommand{\arraystretch}{1}
{\centering\noindent\textbf{Smaller candidate noise}\par}\nobreak\vspace{1pt}
\noindent\begin{tabular}{@{}lrrr@{}}\toprule
$\sigma$ & Value & Accuracy \% & Gain pp, simultaneous 95\% CI \\\midrule
0.3 & MLP probe & 43.11 & 0.78 [-0.94,2.50] \\
0.3 & \clr{} & 42.44 & 0.11 [-1.23,1.45] \\
0.75 & MLP probe & 42.78 & 0.44 [-2.51,3.40] \\
0.75 & \clr{} & 41.67 & -0.67 [-3.22,1.89] \\
\bottomrule\end{tabular}
\par\endgroup\vspace{3pt}
\end{minipage}\par
Voting is majority voting over independent continuations. The oracle reads gold labels and is charged for its continuations; it marks the headroom available to a perfect value function. Intervals in the 90-setting grid are simultaneous over all settings. The best setting, the MLP probe with beam search at $N=4$, $b=4$, $h=6$, reaches 41.44\% accuracy, $-0.89$ pp from greedy with simultaneous interval $[-4.69,2.92]$ pp. The four smaller-noise comparisons reuse the frozen readers with $N=8$, $b=4$ and branching at every step, and do not touch the test set.\par
\endgroup\twocolumn\begingroup\fontsize{8.5}{9.8}\selectfont
\section{Audit generator and measurements}

\par\vspace{3pt}\begingroup\fontsize{8}{9}\selectfont\setlength{\tabcolsep}{2.8pt}\renewcommand{\arraystretch}{1}
{\centering\noindent\textbf{Generator and representation contract}\par}\nobreak\vspace{1pt}
\noindent\begin{tabularx}{\linewidth}{@{}lX@{}}\toprule
Item & Specification \\\midrule
Generator & Synthetic arithmetic and graph episodes combine public task variables with one private team fact and a public route permutation. Counterfactuals swap that private semantic fact on supporting records, preserving public variables and nuisance records. Mismatches use another group in the same split. \\
Messages & Qwen3-4B sender and receiver share a fixed checkpoint. Full-prefill post-RoPE keys and values retain physical positions, masks and original cohort padding. \\
Commitments & \texttt{COMMITMENT: choose Route \{route\}.}\newline\texttt{COMMITMENT: select Route \{route\}.}\newline\texttt{COMMITMENT: the conclusion is Route \{route\}.} \\
Feature map & Layers 7/16/25/34; record means, public-prefix mean and last 8 valid tokens. Concatenated K/V are projected from 2048 to 512 dimensions with fixed orthogonal columns. \\
Panel / calibration & Train / reader-val / $\mathrm{cal}_A$ / $\mathrm{cal}_B$ / test: 8000 / 1000 / 500 / 1000 / 2000 groups. Taskwise temperature then clean conformal thresholds; questions derive expectations from public variables and commitment. \\
\bottomrule\end{tabularx}
\par\endgroup\vspace{3pt}
\par\vspace{3pt}\begingroup\centering\fontsize{7.5}{8.4}\selectfont\setlength{\tabcolsep}{2.8pt}\renewcommand{\arraystretch}{1}
{\centering\noindent\textbf{Audit test metrics per sub-question, seen templates unless marked held-out}\par}\nobreak\vspace{1pt}
\noindent\begin{tabular}{@{}lrrrrr@{}}\toprule
Reader & Metric & A1 fact & A2 support & A3 consist. & A4 private \\\midrule
\clr{} & Accuracy & 0.995 & 0.500 & 0.510 & 0.993 \\
 & NLL & 0.037 & 0.693 & 0.693 & 0.020 \\
 & Brier & 0.007 & 0.500 & 0.499 & 0.010 \\
 & ECE & 0.002 & 0.001 & 0.004 & 0.003 \\
 & AUROC & 0.999 & 0.500 & 0.519 & NA \\
 & Cov. .05 & 0.963 & 0.943 & 0.961 & 0.950 \\
 & Size .05 & 0.964 & 1.886 & 1.915 & 0.951 \\
 & Cov. .01 & 0.993 & 0.989 & 0.989 & 0.990 \\
 & Size .01 & 0.995 & 1.978 & 1.976 & 0.994 \\
 & Held-out acc. & 0.884 & 0.500 & 0.511 & 0.993 \\
MLP probe & Accuracy & 0.683 & 0.498 & 0.497 & 0.455 \\
 & NLL & 0.559 & 0.693 & 0.693 & 1.169 \\
 & Brier & 0.386 & 0.500 & 0.500 & 0.635 \\
 & ECE & 0.017 & 0.008 & 0.012 & 0.041 \\
 & AUROC & 0.770 & 0.499 & 0.498 & NA \\
 & Cov. .05 & 0.958 & 0.944 & 0.957 & 0.947 \\
 & Size .05 & 1.663 & 1.888 & 1.917 & 3.143 \\
 & Cov. .01 & 0.990 & 0.989 & 0.990 & 0.987 \\
 & Size .01 & 1.815 & 1.977 & 1.981 & 3.556 \\
 & Held-out acc. & 0.683 & 0.498 & 0.497 & 0.455 \\
Linear probe & Accuracy & 0.670 & 0.496 & 0.498 & 0.462 \\
 & NLL & 0.592 & 0.694 & 0.694 & 1.178 \\
 & Brier & 0.410 & 0.501 & 0.501 & 0.643 \\
 & ECE & 0.014 & 0.019 & 0.019 & 0.029 \\
 & AUROC & 0.742 & 0.497 & 0.498 & NA \\
 & Cov. .05 & 0.952 & 0.952 & 0.949 & 0.950 \\
 & Size .05 & 1.709 & 1.906 & 1.898 & 3.204 \\
 & Cov. .01 & 0.988 & 0.991 & 0.990 & 0.992 \\
 & Size .01 & 1.867 & 1.981 & 1.981 & 3.658 \\
 & Held-out acc. & 0.670 & 0.496 & 0.498 & 0.462 \\
Question only & Accuracy & 0.500 & 0.500 & 0.500 & 0.253 \\
 & NLL & 0.693 & 0.693 & 0.694 & 1.386 \\
 & Brier & 0.500 & 0.500 & 0.501 & 0.750 \\
 & ECE & 0.002 & 0.002 & 0.016 & 0.003 \\
 & AUROC & 0.500 & 0.500 & 0.500 & NA \\
 & Cov. .05 & 1.000 & 0.978 & 0.833 & 1.000 \\
 & Size .05 & 2.000 & 1.957 & 1.667 & 4.000 \\
 & Cov. .01 & 1.000 & 1.000 & 0.833 & 1.000 \\
 & Size .01 & 2.000 & 2.000 & 1.667 & 4.000 \\
 & Held-out acc. & 0.500 & 0.500 & 0.500 & 0.254 \\
\bottomrule\end{tabular}
\par\endgroup\vspace{3pt}
A1 asks for the private fact, A2 whether the message supports the route, A3 whether it is consistent with the commitment, and A4 which team it refers to. AUROC is undefined for A4. Seen and held-out templates share one calibrator. The source CSVs also contain 95\% intervals from 10000 group resamples, conditional on the frozen seeds and calibrators.\par
\par\vspace{3pt}\begingroup\centering\fontsize{8}{9}\selectfont\setlength{\tabcolsep}{2.8pt}\renewcommand{\arraystretch}{1}
{\centering\noindent\textbf{CF detection at nominal 5\% FPR}\par}\nobreak\vspace{1pt}
\noindent\begin{tabular}{@{}lrrrr@{}}\toprule
Method & $k$ & Detect \% & FPR \% & AUROC \\\midrule
\clr{} & 1 & 28.7 & 3.35 & 0.789 \\
 & 4 & 72.8 & 5.47 & 0.928 \\
 & 16 & 93.1 & 5.07 & 0.986 \\
MLP probe & 1 & 14.5 & 5.28 & 0.589 \\
 & 4 & 20.5 & 5.18 & 0.679 \\
 & 16 & 23.9 & 4.22 & 0.735 \\
Linear probe & 1 & 14.6 & 5.32 & 0.582 \\
 & 4 & 19.7 & 5.42 & 0.678 \\
 & 16 & 21.8 & 4.72 & 0.733 \\
Question only & 1 & 1.7 & 1.68 & 0.500 \\
 & 4 & 4.7 & 4.68 & 0.500 \\
 & 16 & 4.3 & 4.33 & 0.500 \\
Clean/CF classifier & NA & 4.9 & 4.95 & 0.500 \\
KV norm & NA & 6.5 & 6.80 & 0.500 \\
Text only & NA & 0.0 & 0.00 & 0.533 \\
\bottomrule\end{tabular}
\par\endgroup\vspace{3pt}
\par\vspace{3pt}\begingroup\centering\fontsize{7.5}{8.4}\selectfont\setlength{\tabcolsep}{2.8pt}\renewcommand{\arraystretch}{1}
{\centering\noindent\textbf{Attention-probe reader, sub-question accuracy in \%}\par}\nobreak\vspace{1pt}
\noindent\begin{tabular}{@{}lrrr@{}}\toprule
Split & Reader & A1 fact & A4 private value \\\midrule
reader-val & \clr{} & 99.7 [99.6,99.8] & 99.5 [99.3,99.7] \\
reader-val & Attention probe & 70.6 [70.0,71.2] & 97.8 [97.3,98.3] \\
test & \clr{} & 99.5 [99.4,99.7] & 99.3 [99.1,99.5] \\
test & Attention probe & 71.0 [70.6,71.5] & 98.0 [97.7,98.3] \\
\bottomrule\end{tabular}
\par\endgroup\vspace{3pt}
\par{\centering\noindent\includegraphics[width=\linewidth]{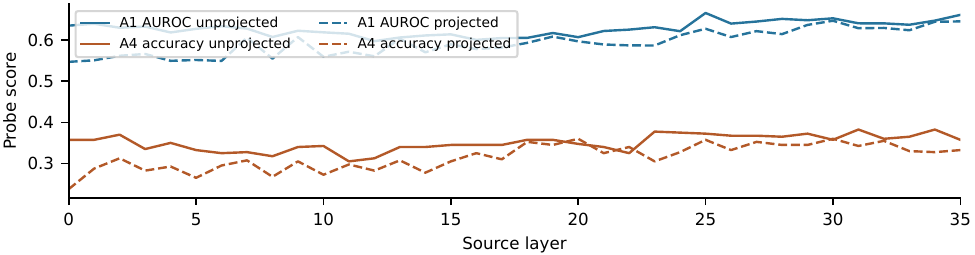}\par}
The layer scan fits probes on 800 training groups and scores them on 200 held-out training groups; the test set is not used. Layer 25 maximizes the combined probe criterion, and the retained layers are those at offsets of 9 from it, modulo 36. The projection preserves A1 accuracy exactly and costs 1.75 pp on A4.\par
\par\vspace{3pt}\begingroup\centering\fontsize{7.5}{8.4}\selectfont\setlength{\tabcolsep}{2.8pt}\renewcommand{\arraystretch}{1}
{\centering\noindent\textbf{Natural A3 consistency probability}\par}\nobreak\vspace{1pt}
\noindent\begin{tabular}{@{}lrrrr@{}}\toprule
Split / mode & Correct? & $n$ & Mean & 10--90\% \\\midrule
val/CoT & No & 49 & 0.497 & 0.488--0.506 \\
val/CoT & Yes & 51 & 0.496 & 0.489--0.502 \\
val/direct & No & 67 & 0.497 & 0.490--0.506 \\
val/direct & Yes & 33 & 0.495 & 0.487--0.503 \\
test/CoT & No & 184 & 0.499 & 0.491--0.507 \\
test/CoT & Yes & 116 & 0.498 & 0.489--0.507 \\
test/direct & No & 232 & 0.499 & 0.490--0.507 \\
test/direct & Yes & 68 & 0.498 & 0.489--0.507 \\
\bottomrule\end{tabular}
\par\endgroup\vspace{3pt}
\par\vspace{3pt}\begingroup\centering\fontsize{7.5}{8.4}\selectfont\setlength{\tabcolsep}{2.8pt}\renewcommand{\arraystretch}{1}
{\centering\noindent\textbf{HiddenBench test, six independent groups}\par}\nobreak\vspace{1pt}
\noindent\begin{tabular}{@{}lrrr@{}}\toprule
Reader & A1 fact \% & A4 private \% & Mismatch \% \\\midrule
\clr{} & 50.9 [50.3,51.6] & 43.6 [34.0,53.0] & 0.0 [0.0,0.0] \\
MLP probe & 50.7 [49.3,52.4] & 47.2 [41.0,54.9] & 0.0 [0.0,0.0] \\
Attention probe & 55.2 [53.5,57.6] & 38.9 [28.5,47.2] & 0.0 [0.0,0.0] \\
\bottomrule\end{tabular}
\par\endgroup\vspace{3pt}
The $k$ ablations take prefixes of a single 16-question sequence, so results for different $k$ are correlated. The natural direct-answer and chain-of-thought (CoT, 512 tokens) panels have val 100 and test 300. HiddenBench provides 65 cases in 63 groups, split 13/20/4/20/6, and admits only mismatch tests. With 20 $\mathrm{cal}_B$ groups the 1\% conformal threshold is infinite by construction, so no alarm can fire; at this scale HiddenBench serves as a format check rather than a test of detection power.\par
\endgroup\clearpage\begingroup\fontsize{8}{9.2}\selectfont
\section{Robustness and perturbation search}

\par\vspace{3pt}\begingroup\fontsize{8}{9}\selectfont\setlength{\tabcolsep}{2.8pt}\renewcommand{\arraystretch}{1}
{\centering\noindent\textbf{Threat and optimization protocol}\par}\nobreak\vspace{1pt}
\noindent\begin{tabularx}{\linewidth}{@{}lX@{}}\toprule
Item & Setting \\\midrule
Threat & Change transmitted raw K/V while public input and commitment remain fixed; relative per-layer joint K/V norm budgets apply after BF16 projection. \\
Non-adaptive & CF substitution, noise, scale, sign, norm-matched replacement and adapted DiffMean controls. \\
Unread-layer / null-space & The first changes layers outside the original reader; the second uses equal-and-opposite token-pair updates in a record, preserving its pooled feature and stored fp16 representation. \\
White-box / transfer & The white-box attack maximizes receiver target log probability minus weighted expected audit NLL; the transfer attack moves attacks between independently initialized reader seeds. \\
PGD & Zero initialization; budgets 50/200 iterations; per-layer step size $\epsilon/4$; BF16 feasible projection with straight-through gradients; early success may reuse a perturbation at both budgets. \\
EOT & The pooled-auditor controls rotate 4 independent panels; the matched matrix uses fresh EOT16 and 2 independent selection panels. Final report questions never enter optimization or stopping. \\
Auditors & Pooled: fixed pooling, as in the main paper. Randomized: random 8 of 36 layers, boundary jitter and 128 dimensions. All-layer: all 36 layers at 128 dimensions. Receiver state: receiver states at 512 dimensions. \\
\bottomrule\end{tabularx}
\par\endgroup\vspace{3pt}
\par\vspace{3pt}\begingroup\centering\fontsize{7.5}{8.4}\selectfont\setlength{\tabcolsep}{2.8pt}\renewcommand{\arraystretch}{1}
{\centering\noindent\textbf{Failure percentage at $\epsilon=.03$ by attack and auditor}\par}\nobreak\vspace{1pt}
\noindent\begin{tabular}{@{}lrrrr@{}}\toprule
Attack & Pooled & Randomized & All-layer & Receiver \\\midrule
Non-adaptive & --$^{a}$ & -- & -- & -- \\
Unread-layer & \shortstack[r]{69.9\\{[66.0,73.5]}} & \shortstack[r]{70.1\\{[66.3,73.8]}} & \shortstack[r]{69.1\\{[65.1,72.8]}} & \shortstack[r]{33.3\\{[29.7,36.9]}} \\
Null-space & \shortstack[r]{69.9\\{[66.0,73.5]}} & \shortstack[r]{69.4\\{[65.5,73.1]}} & \shortstack[r]{69.1\\{[65.1,72.8]}} & \shortstack[r]{57.7\\{[53.9,61.6]}} \\
White-box & \shortstack[r]{70.3\\{[66.4,73.9]}} & \shortstack[r]{70.6\\{[66.7,74.3]}} & \shortstack[r]{71.8\\{[67.9,75.5]}} & \shortstack[r]{71.1\\{[67.1,74.7]}} \\
Transfer & --$^{a}$ & -- & -- & -- \\
\bottomrule\end{tabular}
\par\endgroup\vspace{3pt}
Dashes mark cells outside the matched matrix; superscript $a$ points to the pooled-auditor controls in the next table, which use all four question types and are therefore reported separately from the A1/A4 matrix. A failure is an episode where the answer of the receiver moves from a non-target answer to the CF target and the auditor raises no alarm; rates are over all 500 groups. All cells use $k=16$, $\lambda=1$, a 200-step budget and nominal 5\% FPR.\par
\par\vspace{3pt}\begingroup\centering\fontsize{8}{9}\selectfont\setlength{\tabcolsep}{2.8pt}\renewcommand{\arraystretch}{1}
{\centering\noindent\textbf{Non-adaptive and transfer controls against the pooled auditor}\par}\nobreak\vspace{1pt}
\noindent\begin{tabular}{@{}lr@{}}\toprule
Control & Failure \%, 95\% interval \\\midrule
CF substitution & 0.7 [0.3,1.2] \\
Noise & 2.0 [0.9,3.3] \\
Scale $\times8$ & 16.5 [13.5,19.5] \\
Sign flip & 7.7 [5.9,9.4] \\
Norm-matched random & 9.6 [7.7,11.5] \\
DiffMean & 0.8 [0.2,1.6] \\
Transfer attack & 68.9 [65.0,72.5] \\
\bottomrule\end{tabular}
\par\endgroup\vspace{3pt}
\par{\centering\noindent\includegraphics[width=\linewidth]{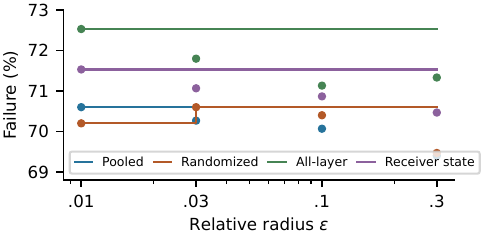}\par}
Circles are individual frozen PGD results, and the step curve is their running maximum over smaller radii. The curve makes finite-search point estimates monotone in $\epsilon$; like any finite PGD search, it lower-bounds the worst case. The randomized, all-layer and receiver-state auditors share one adapter-training protocol; the pooled auditor reuses its earlier body.\par
\par\vspace{3pt}\begingroup\centering\fontsize{7.5}{8.4}\selectfont\setlength{\tabcolsep}{2.8pt}\renewcommand{\arraystretch}{1}
{\centering\noindent\textbf{Message-dependent receiver: white-box failure percentage}\par}\nobreak\vspace{1pt}
\noindent\begin{tabular}{@{}lrr@{}}\toprule
$\epsilon$ & Message auditor & Receiver-state auditor \\\midrule
0.01 & 96.5 [94.0,98.5] & 81.7 [78.3,85.0] \\
0.1 & 96.0 [93.5,98.2] & 72.0 [68.3,75.5] \\
1.0 & 95.3 [92.7,97.5] & 64.0 [60.3,67.7] \\
\bottomrule\end{tabular}
\par\endgroup\vspace{3pt}
\par\vspace{3pt}\begingroup\centering\fontsize{7.5}{8.4}\selectfont\setlength{\tabcolsep}{2.8pt}\renewcommand{\arraystretch}{1}
{\centering\noindent\textbf{Minimum-search ratio, auditor-constrained / receiver-only}\par}\nobreak\vspace{1pt}
\noindent\begin{tabular}{@{}lrrr@{}}\toprule
Reader & Finite pairs & q10 / q50 / q90 & Accept \% \\\midrule
Message & 45 & 1.0/1.0/1.0 & 95.6 \\
Receiver state & 45 & 1.0/1.0/2.0 & 93.3 \\
\bottomrule\end{tabular}
\par\endgroup\vspace{3pt}
The primary test has 200 groups and three seeds. On the 200-group dependency check, receiver accuracy with the true message, no message and a mismatched message is 93/27/26\%. The minimum-radius search covers 15 groups, giving 45 seed pairs per design; both searches return a finite radius in all 45 pairs, with no pair censored or already at the target. At most 6 bisections locate each search interval.\par
\section{Computational cost}

\par\vspace{3pt}\begingroup\centering\fontsize{7.5}{8.4}\selectfont\setlength{\tabcolsep}{2.8pt}\renewcommand{\arraystretch}{1}
{\centering\noindent\textbf{Compute by experiment block}\par}\nobreak\vspace{1pt}
\noindent\begin{tabular}{@{}lrrr@{}}\toprule
Experiment & GPUh & Source calls & Missing rows \\\midrule
Source replay & NA & NA & 1 \\
Value labels & $\ge$23.67 & 2855126 & 0 \\
GPT-2 noise scan & $\ge$36.30 & $\ge$3506255 & 2 \\
Reader training & $\ge$53.81 & 117826 & 0 \\
Questions and probes & $\ge$2.40 & 200689 & 0 \\
Reader diagnostics & 34.63 & $\ge$0 & 1 \\
Search, compute-matched & 5.81 & $\ge$1124771 & 1 \\
Search grid and siblings & 16.81 & 1729285 & 0 \\
Cross-source, LLaMA scan & 47.60 & 1907392 & 0 \\
Auditing, pilot & 7.73 & 59274 & 0 \\
Auditing, full panel & 15.94 & 126485 & 0 \\
Robustness controls & 9.57 & 122695 & 0 \\
Robustness matrix & 19.17 & $\ge$157833 & 4 \\
Dependent receiver & 20.97 & $\ge$159655 & 2 \\
Joint training & 7.38 & 0 & 0 \\
Final comparisons & 13.52 & 202 & 0 \\
Verbalizer & 17.25 & 0 & 0 \\
\bottomrule\end{tabular}
\par\endgroup\vspace{3pt}
GPUh is the summed resident wall time of processes on RTX 4090 workers, including loading and restarts; entries marked $\ge$ are lower bounds where logs are incomplete. Source calls are physical batched calls, distinct from the per-branch equivalents used as the search cost axis. Reused training is charged to the block that first ran it.\par
\par\vspace{3pt}\begingroup\centering\fontsize{7.5}{8.4}\selectfont\setlength{\tabcolsep}{2.8pt}\renewcommand{\arraystretch}{1}
{\centering\noindent\textbf{Accuracy and cost at batches 1 / 64}\par}\nobreak\vspace{1pt}
\noindent\begin{tabular}{@{}lrrr@{}}\toprule
Type / reader & Acc. \% & GFLOPs/query & ms/batch \\\midrule
Q3 / MLP probe & 58.6 & 0.000861/0.000861 & 0.58/1.49 \\
Q3 / Attention probe & 65.4 & 0.00726/0.00726 & 1.72/3.37 \\
Q3 / \clr{} & 76.1 & 0.254/0.311 & 7.79/10.06 \\
Q3 / Verbalizer & 76.8 & 165/193 & 151.97/802.07 \\
Q4 / MLP probe & 68.3 & 0.000215/0.000215 & 0.52/1.63 \\
Q4 / Attention probe & 72.0 & 0.00525/0.00525 & 1.83/5.40 \\
Q4 / \clr{} & 76.2 & 0.175/0.256 & 7.69/8.87 \\
Q4 / Verbalizer & 76.3 & 76.6/121 & 144.86/392.90 \\
\bottomrule\end{tabular}
\par\endgroup\vspace{3pt}
\par\vspace{3pt}\begingroup\centering\fontsize{7.5}{8.4}\selectfont\setlength{\tabcolsep}{2.8pt}\renewcommand{\arraystretch}{1}
{\centering\noindent\textbf{Training time per reader}\par}\nobreak\vspace{1pt}
\noindent\begin{tabular}{@{}lrr@{}}\toprule
Reader & Total GPUh & Seed 11 / 22 / 33 GPUh \\\midrule
\clr{} & 6.030 & 1.944/2.118/1.967 \\
Attention probe & 2.353 & 0.815/0.783/0.756 \\
Verbalizer & 16.719 & 5.555/6.879/4.284 \\
MLP probe & NA & Not timed separately \\
\bottomrule\end{tabular}
\par\endgroup\vspace{3pt}
\endgroup

\section{Conformal set sizes}
\label{app:cp}
\Cref{tab:cp} gives the conformal set sizes and coverage summarized in the main text.
\begin{table}[t]
\centering
\small
\setlength{\tabcolsep}{3pt}
\begin{tabular}{lcccc}
\toprule
& Q1 value & Q2 exit & Q3 answer & Q4 interm.\\
\midrule
MLP probe & 1.41 / 90.5 & 1.33 / 89.0 & 2.16 / 92.0 & 1.53 / 90.1\\
MLP, prefix & 1.37 / 89.4 & 1.30 / 88.9 & 2.05 / 92.4 & 1.56 / 91.2\\
\clr{} & 1.44 / 90.6 & 1.37 / 89.7 & \textbf{1.49} / 91.6 & \textbf{1.35} / 89.7\\
\bottomrule
\end{tabular}
\caption{Mondrian split conformal at $\alpha=0.1$ on CODI-GPT-2 test: mean set size / empirical coverage in \%. The 1.7B verbalizer, evaluated on Q3 and Q4, covers 92.1\% and 90.2\% with sets within 0.02 of \clr{}'s.}
\label{tab:cp}
\end{table}

\section{Additional details for the main text}
\label{app:details}
\paragraph{Baseline adaptations.} Our HSRM-style reader applies a two-layer Transformer encoder of width 256 with four heads to the frozen latent prefix, followed by masked mean pooling, layer normalization and a scalar head. It is trained on Q1 trajectory outcomes with a Bernoulli loss, whereas the original uses a pairwise ranking loss over explicit chains of thought, and separate temperatures for Q1 and Q2 are fitted on $\mathrm{cal}_A$. CLUE represents a prefix by $z_t-z_0$ in the last layer, whereas the original uses all-layer deltas of an explicit chain of thought; it compares the Euclidean distances to success and failure centroids from complete reader-training trajectories and maps the distance difference to a probability by logistic regression on $\mathrm{cal}_A$.

\paragraph{Option-order equivariance.} Without the block mask, predictions change under option permutation for 0.29\% of questions, against 0.14\% caused by numerical ties with the mask.

\paragraph{Conformal sets on Q1 and Q2.} On the state-only questions, the set sizes of \clr{} and the MLP probes lie within 0.07 of each other (\cref{tab:cp}).

\paragraph{Weighted conformal under shift.} Standard Mondrian conformal is the better choice under our shifts: weighted split conformal prediction with an estimated density ratio does not reduce under-coverage and, under step shift, where the supports of early and late states barely overlap, over-covers with sets about twice as large. Of the 32 shift settings, 6 under-cover the nominal 90\% by more than 2 points (none by more than 5); all involve dataset, source or step shift.

\paragraph{Latent dynamics of CODI.} Early-exit accuracy rises mainly after odd steps (\cref{fig:search}a), consistent with CODI's alternation between value and operator thoughts. Perturbing only steps 4 to 6 costs 5 points, and the fraction of states with non-degenerate value falls from 41\% at $t=0$ to 10\% at $t=5$.

\paragraph{Relay receiver.} In the LatentMAS relay setting, the message raises receiver accuracy from 23.0 to 28.7\%; the receiver is correct on 30.8\% of clean messages in the robustness panel.

\section{Latent embeddings}
\label{app:qual}
\begin{figure}[t]
\centering
\includegraphics[width=\linewidth]{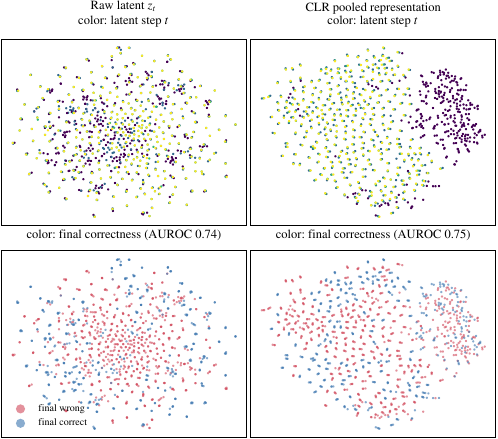}
\caption{t-SNE of CODI-GPT-2 test latents and of \clr{}'s pooled latent representation, colored by latent step and by final correctness. AUROC is five-fold logistic regression for final correctness on the plotted vectors. \clr{}'s representation separates the initial state from later steps; final correctness is read out by the trained heads rather than visible in the two-dimensional geometry.}
\label{fig:qual-reasoning}
\end{figure}

\Cref{fig:qual-reasoning} shows t-SNE maps of clean CODI-GPT-2 test latents for 300 problems, selected by a fixed hash order, at $t\in\{0,2,4,6\}$.

\section{Per-problem Q4 readouts}
\label{app:qual-reading}
\begin{figure*}[t]
\centering
\includegraphics[width=\textwidth]{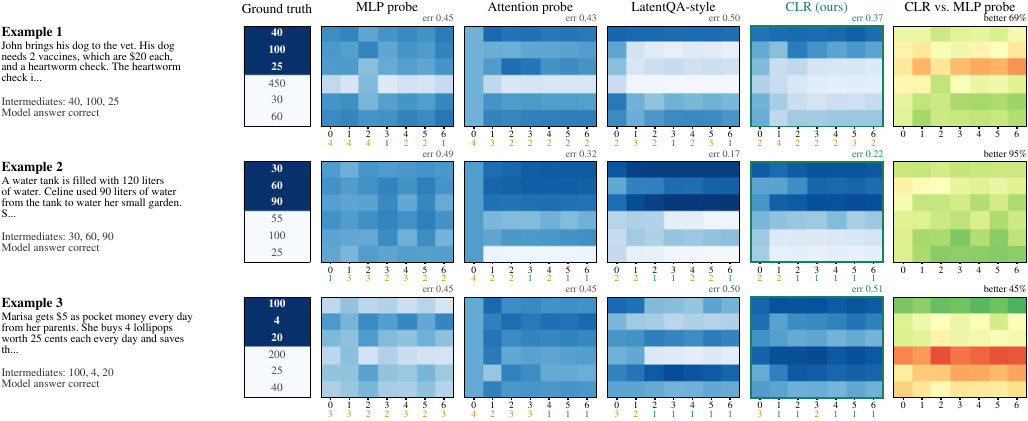}
\caption{Q4 readouts for the first three selected CODI-GPT-2 test problems.}
\label{fig:qual-reading-main}
\end{figure*}

We select test problems whose reference solution has at least three intermediate values and take them in a fixed hash order of their IDs, without looking at any model output; we show the first twelve for each source. Each row is one problem. The left column gives the problem text and its intermediate values. The ground-truth column lists the candidate values of Q4, the true intermediates on a dark background and the distractors, which are intermediates of other problems, on a light background. Each heatmap shows one reader's $P(\text{yes})$ that the candidate in that row is an intermediate value, for latent steps $t=0,\dots,6$ along the horizontal axis; a perfect reader reproduces the ground-truth column at every step. ``err'' is the mean absolute error against the ground truth. The digits below a heatmap give the size of the $\alpha=0.1$ conformal set for Q3 at each step: green for a single correct option, amber for a larger set that contains the model's answer, red for a set that misses it. The last column compares \clr{} with the MLP probe and is green where \clr{} is closer to the ground truth. \Cref{fig:qual-reading-main} shows the first three GPT-2 problems at a larger scale. The third example shows the typical error mode: values that occur in the problem text but are not intermediate results (200 and 25) receive high probability.

\begin{figure*}[p]
\centering
\includegraphics[width=\textwidth,height=0.92\textheight,keepaspectratio]{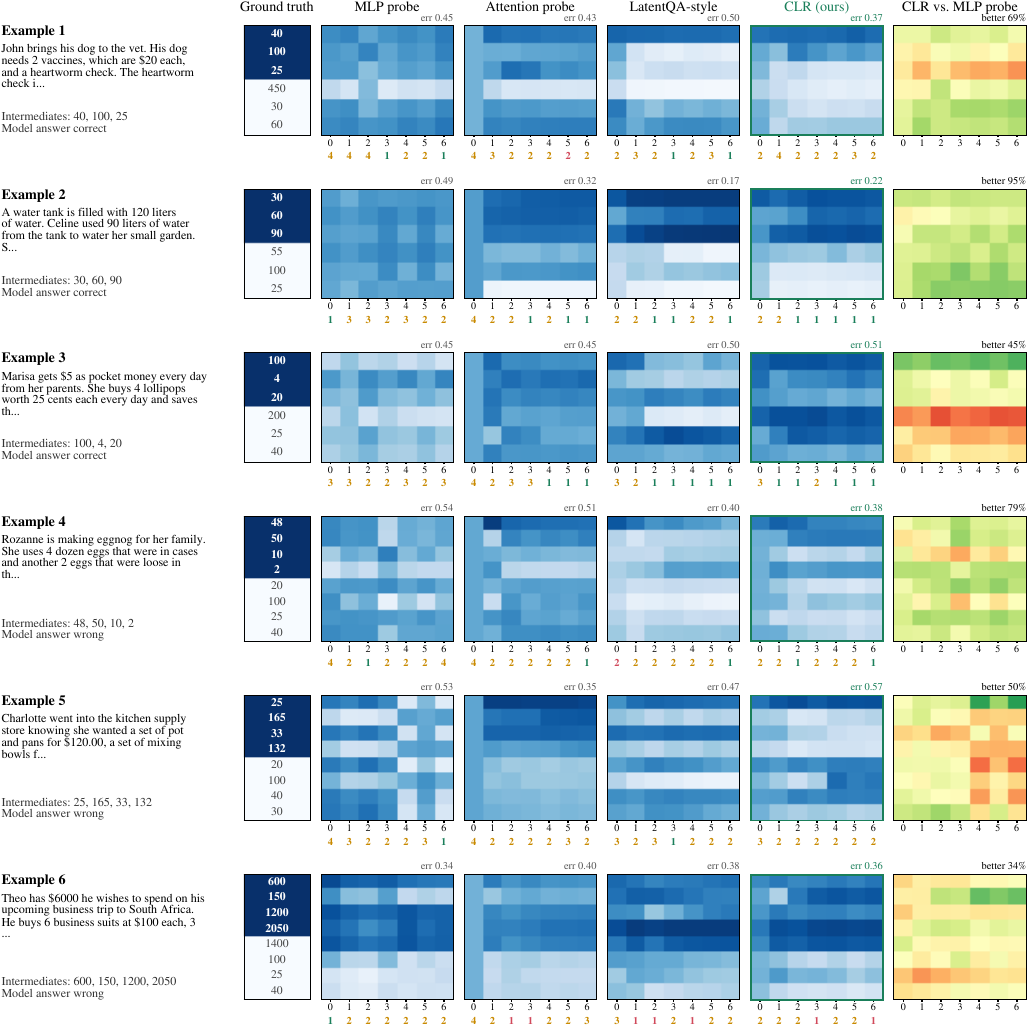}
\caption{Q4 readouts and Q3 conformal set sizes for selected CODI-GPT-2 test problems 1--6.}
\end{figure*}
\begin{figure*}[p]
\centering
\includegraphics[width=\textwidth,height=0.92\textheight,keepaspectratio]{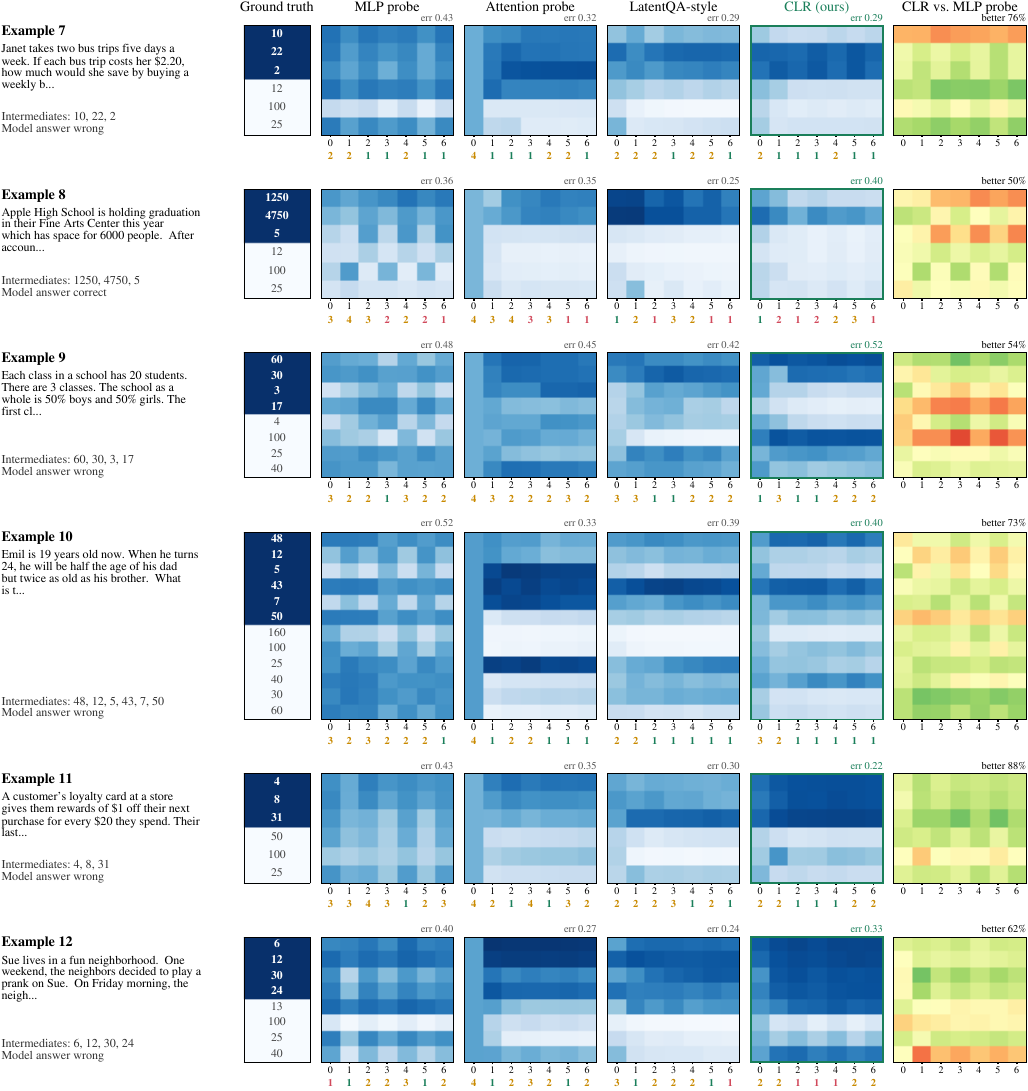}
\caption{Q4 readouts and Q3 conformal set sizes for selected CODI-GPT-2 test problems 7--12.}
\end{figure*}
\begin{figure*}[p]
\centering
\includegraphics[width=\textwidth,height=0.92\textheight,keepaspectratio]{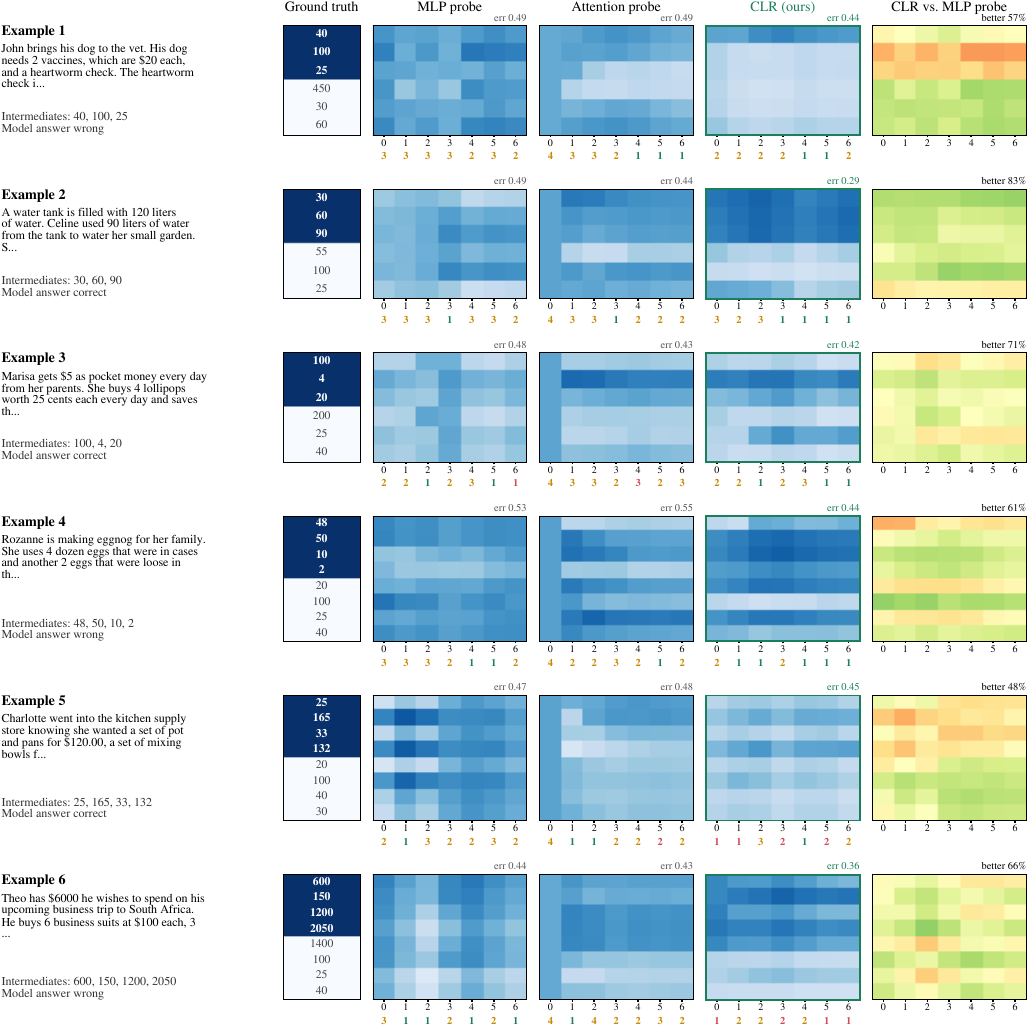}
\caption{Q4 readouts and Q3 conformal set sizes for selected CODI-LLaMA test problems 1--6. The LatentQA-style verbalizer is evaluated on GPT-2 only, so it has no column.}
\end{figure*}
\begin{figure*}[p]
\centering
\includegraphics[width=\textwidth,height=0.92\textheight,keepaspectratio]{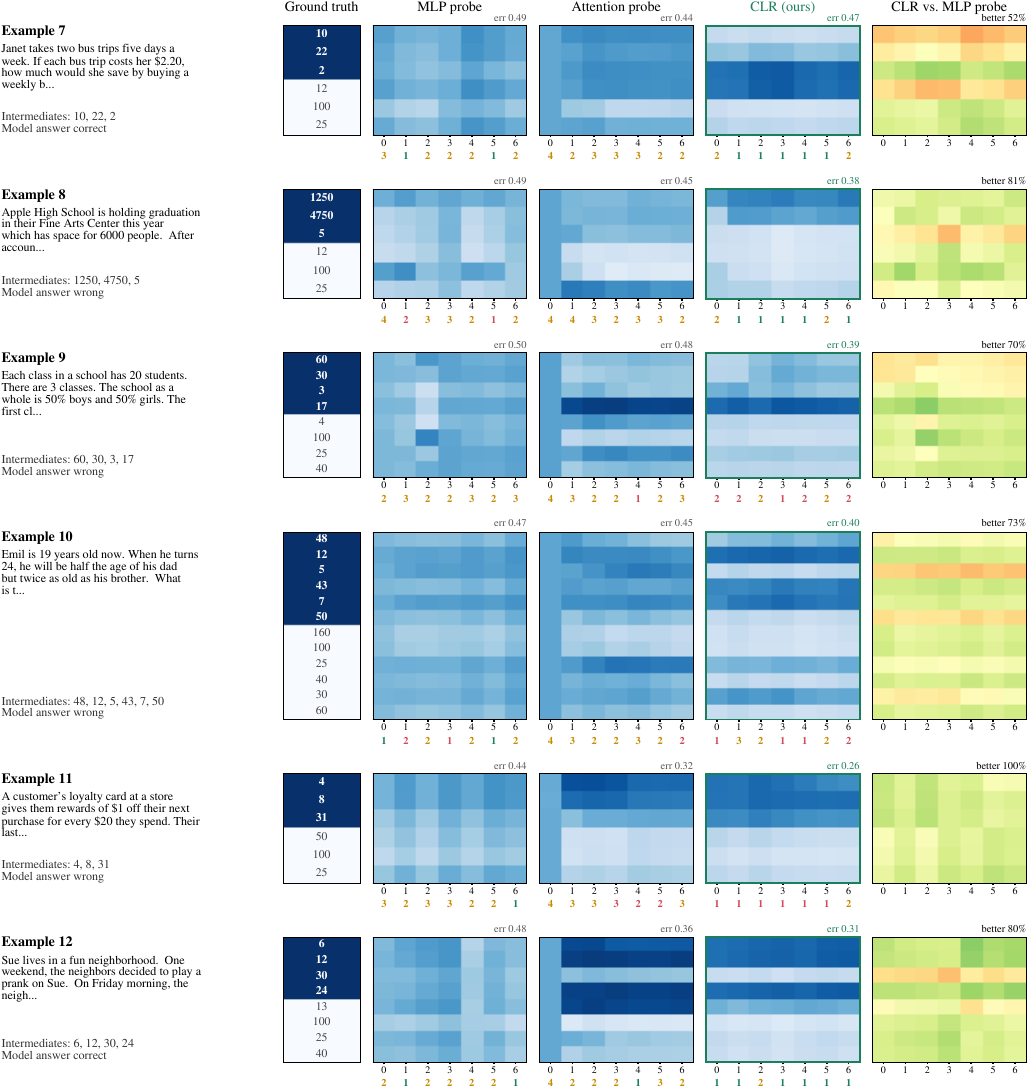}
\caption{Q4 readouts and Q3 conformal set sizes for selected CODI-LLaMA test problems 7--12.}
\end{figure*}


{\small
\begin{thebibliography}{51}
\providecommand{\natexlab}[1]{#1}
\providecommand{\url}[1]{\texttt{#1}}
\expandafter\ifx\csname urlstyle\endcsname\relax
  \providecommand{\doi}[1]{doi: #1}\else
  \providecommand{\doi}{doi: \begingroup \urlstyle{rm}\Url}\fi

\bibitem[Alain and Bengio(2016)]{alain2016understanding}
Guillaume Alain and Yoshua Bengio.
\newblock Understanding intermediate layers using linear classifier probes.
\newblock \emph{arXiv preprint arXiv:1610.01644}, 2016.

\bibitem[Angelopoulos and
  Bates(2023)]{angelopoulos2022gentleintroductionconformalprediction}
Anastasios~N. Angelopoulos and Stephen Bates.
\newblock Conformal prediction: A gentle introduction.
\newblock \emph{Foundations and Trends in Machine Learning}, 16\penalty0
  (4):\penalty0 494--591, 2023.

\bibitem[Antunes et~al.(2026)Antunes, Albuquerque, Schmith, Batista, da~Silva,
  Rossi, da~Silva, da~Costa, Bertotto, and Fausto]{antunes2026narrow}
Carlos~Andr{\'e} Antunes, Aldry Albuquerque, Gabriel de~Melo Schmith, Nat{\~a}
  Batista, Pedro Oziel~Escobar da Silva, Stefano~Benko Rossi, Lingrid
  Dietrich~Lorenz da Silva, Fl{\'a}via~Cristina da Costa, Thiago Bertotto, and
  Luis~Fernando Fausto.
\newblock Narrow decision models: A {System-1} architecture for calibrated,
  typed decisions.
\newblock \emph{arXiv preprint arXiv:2610.25553}, 2026.

\bibitem[Asif et~al.(2026)Asif, Amiri, Abbas, Sattigeri, and
  Ramamurthy]{asif2026lcguard}
Sadia Asif, Mohammad~Mohammadi Amiri, Momin Abbas, Prasanna Sattigeri, and
  Karthikeyan~Natesan Ramamurthy.
\newblock {LCGuard}: Latent communication guard for safe {KV} sharing in
  multi-agent systems.
\newblock \emph{arXiv preprint arXiv:2605.22786}, 2026.

\bibitem[Athalye et~al.(2018)Athalye, Engstrom, Ilyas, and
  Kwok]{athalye2018synthesizing}
Anish Athalye, Logan Engstrom, Andrew Ilyas, and Kevin Kwok.
\newblock Synthesizing robust adversarial examples.
\newblock In \emph{Proceedings of the 35th International Conference on Machine
  Learning}, pages 284--293. PMLR, 2018.

\bibitem[Babu et~al.(2026)Babu, Kaur, Pramanik, Kotevska, Bastian, Jha, Raj,
  Wu, Jha, and Roy]{babu2026consilience}
Abhijith Babu, Ramneet Kaur, Vishal Pramanik, Olivera Kotevska, Nathaniel~D
  Bastian, Susmit Jha, Sunny Raj, Yanzhao Wu, Sumit~Kumar Jha, and Anirban Roy.
\newblock Consilience: Conformally calibrated communication control for
  hidden-profile multi-agent reasoning.
\newblock \emph{arXiv preprint arXiv:2608.20564}, 2026.

\bibitem[Baker et~al.(2025)Baker, Huizinga, Gao, Dou, Guan, Madry, Zaremba,
  Pachocki, and Farhi]{baker2025monitoring}
Bowen Baker, Joost Huizinga, Leo Gao, Zehao Dou, Melody~Y Guan, Aleksander
  Madry, Wojciech Zaremba, Jakub Pachocki, and David Farhi.
\newblock Monitoring reasoning models for misbehavior and the risks of
  promoting obfuscation.
\newblock \emph{arXiv preprint arXiv:2503.11926}, 2025.

\bibitem[Belinkov(2022)]{belinkov2022probing}
Yonatan Belinkov.
\newblock Probing classifiers: Promises, shortcomings, and advances.
\newblock \emph{Computational Linguistics}, 48\penalty0 (1):\penalty0 207--219,
  2022.

\bibitem[Belrose et~al.(2023)Belrose, Ostrovsky, McKinney, Furman, Smith,
  Halawi, Biderman, and Steinhardt]{belrose2023tuned}
Nora Belrose, Igor Ostrovsky, Lev McKinney, Zach Furman, Logan Smith, Danny
  Halawi, Stella Biderman, and Jacob Steinhardt.
\newblock Eliciting latent predictions from transformers with the tuned lens.
\newblock \emph{arXiv preprint arXiv:2303.08112}, 2023.

\bibitem[Brito and Baquero(2026)]{brito2026latent}
Lu{\'\i}s Brito and Carlos Baquero.
\newblock When latent agents lie: {KV}-cache integrity in multi-agent {LLM}
  collaboration.
\newblock \emph{arXiv preprint arXiv:2606.28958}, 2026.

\bibitem[Chen et~al.(2024)Chen, Vondrick, and Mao]{chen2024selfie}
Haozhe Chen, Carl Vondrick, and Chengzhi Mao.
\newblock {SelfIE}: Self-interpretation of large language model embeddings.
\newblock In \emph{Proceedings of the 41st International Conference on Machine
  Learning}, pages 7373--7388. PMLR, 2024.

\bibitem[Cheng et~al.(2026)Cheng, Das, and Ramnath]{cheng2026does}
Jiaming Cheng, Subhransu Das, and Rajiv Ramnath.
\newblock When does latent communication pay? a causal audit of relayed {KV}
  caches in multi-agent {LLMs}.
\newblock \emph{arXiv preprint arXiv:2608.04893}, 2026.

\bibitem[Cobbe et~al.(2021)Cobbe, Kosaraju, Bavarian, Chen, Jun, Kaiser,
  Plappert, Tworek, Hilton, Nakano, Hesse, and Schulman]{cobbe2021training}
Karl Cobbe, Vineet Kosaraju, Mohammad Bavarian, Mark Chen, Heewoo Jun, Lukasz
  Kaiser, Matthias Plappert, Jerry Tworek, Jacob Hilton, Reiichiro Nakano,
  Christopher Hesse, and John Schulman.
\newblock Training verifiers to solve math word problems.
\newblock \emph{arXiv preprint arXiv:2110.14168}, 2021.

\bibitem[Deng et~al.(2023)Deng, Prasad, Fernandez, Smolensky, Chaudhary, and
  Shieber]{deng2023implicit}
Yuntian Deng, Kiran Prasad, Roland Fernandez, Paul Smolensky, Vishrav
  Chaudhary, and Stuart Shieber.
\newblock Implicit chain of thought reasoning via knowledge distillation.
\newblock \emph{arXiv preprint arXiv:2311.01460}, 2023.

\bibitem[Du et~al.(2026)Du, Dong, and Ning]{du2026latent}
Hanwen Du, Yuxin Dong, and Xia Ning.
\newblock Latent thinking optimization: Your latent reasoning language model
  secretly encodes reward signals in its latent thoughts.
\newblock In \emph{International Conference on Learning Representations}, pages
  101000--101032, 2026.

\bibitem[Fu et~al.(2026)Fu, Min, Zhang, Yan, Dai, Ouyang, and
  Wang]{fu2026cache}
Tianyu Fu, Zihan Min, Hanling Zhang, Jichao Yan, Guohao Dai, Wanli Ouyang, and
  Yu Wang.
\newblock Cache-to-cache: Direct semantic communication between large language
  models.
\newblock In \emph{International Conference on Learning Representations}, pages
  43130--43158, 2026.

\bibitem[Geiping et~al.(2025)Geiping, McLeish, Jain, Kirchenbauer, Singh,
  Bartoldson, Kailkhura, Bhatele, and Goldstein]{geiping2026scaling}
Jonas Geiping, Sean McLeish, Neel Jain, John Kirchenbauer, Siddharth Singh,
  Brian Bartoldson, Bhavya Kailkhura, Abhinav Bhatele, and Tom Goldstein.
\newblock Scaling up test-time compute with latent reasoning: A recurrent depth
  approach.
\newblock In \emph{Advances in Neural Information Processing Systems}, pages
  41340--41391, 2025.

\bibitem[Ghandeharioun et~al.(2024)Ghandeharioun, Caciularu, Pearce, Dixon, and
  Geva]{ghandeharioun2024patchscopes}
Asma Ghandeharioun, Avi Caciularu, Adam Pearce, Lucas Dixon, and Mor Geva.
\newblock Patchscopes: A unifying framework for inspecting hidden
  representations of language models.
\newblock In \emph{Proceedings of the 41st International Conference on Machine
  Learning}, pages 15466--15490. PMLR, 2024.

\bibitem[Guo et~al.(2017)Guo, Pleiss, Sun, and Weinberger]{guo2017calibration}
Chuan Guo, Geoff Pleiss, Yu Sun, and Kilian~Q Weinberger.
\newblock On calibration of modern neural networks.
\newblock In \emph{Proceedings of the 34th International Conference on Machine
  Learning}, pages 1321--1330. PMLR, 2017.

\bibitem[Hao et~al.(2025)Hao, Sukhbaatar, Su, Li, Hu, Weston, and
  Tian]{hao2024training}
Shibo Hao, Sainbayar Sukhbaatar, DiJia Su, Xian Li, Zhiting Hu, Jason Weston,
  and Yuandong Tian.
\newblock Training large language models to reason in a continuous latent
  space.
\newblock In \emph{Second Conference on Language Modeling}, 2025.

\bibitem[Hewitt and Liang(2019)]{hewitt2019controls}
John Hewitt and Percy Liang.
\newblock Designing and interpreting probes with control tasks.
\newblock In \emph{Proceedings of the 2019 Conference on Empirical Methods in
  Natural Language Processing and the 9th International Joint Conference on
  Natural Language Processing}, pages 2733--2743, 2019.

\bibitem[Jeon et~al.(2026)Jeon, Jeong, Lee, Cho, and Shin]{jeon2026vision}
Byungwoo Jeon, Yoonwoo Jeong, Hyunseok Lee, Minsu Cho, and Jinwoo Shin.
\newblock Vision-aligned latent reasoning for multi-modal large language model.
\newblock In \emph{Proceedings of the 43rd International Conference on Machine
  Learning}, pages 51151--51168. PMLR, 2026.

\bibitem[Korbak et~al.(2025)Korbak, Balesni, Barnes, Bengio, Benton, Bloom,
  Chen, Cooney, Dafoe, Dragan, et~al.]{korbak2025chain}
Tomek Korbak, Mikita Balesni, Elizabeth Barnes, Yoshua Bengio, Joe Benton,
  Joseph Bloom, Mark Chen, Alan Cooney, Allan Dafoe, Anca Dragan, et~al.
\newblock Chain of thought monitorability: A new and fragile opportunity for
  {AI} safety.
\newblock \emph{arXiv preprint arXiv:2507.11473}, 2025.

\bibitem[Kuhn et~al.(2023)Kuhn, Gal, and Farquhar]{kuhn2023semantic}
Lorenz Kuhn, Yarin Gal, and Sebastian Farquhar.
\newblock Semantic uncertainty: Linguistic invariances for uncertainty
  estimation in natural language generation.
\newblock In \emph{International Conference on Learning Representations}, 2023.

\bibitem[Li et~al.(2026)Li, Ceballos~Arroyo, Rogers, Saphra, and
  Wallace]{li2025activation}
Millicent Li, Alberto~Mario Ceballos~Arroyo, Giordano Rogers, Naomi Saphra, and
  Byron~C Wallace.
\newblock Do activation verbalization methods convey privileged information?
\newblock In \emph{Proceedings of the 43rd International Conference on Machine
  Learning}, pages 66102--66142. PMLR, 2026.

\bibitem[Li and Zhu(2026)]{li2026hsrm}
Xianzhi Li and Xiaodan Zhu.
\newblock {HSRM}: Hidden-state reward models for test-time verification.
\newblock In \emph{Proceedings of the 2026 Conference on Empirical Methods in
  Natural Language Processing}, pages 39947--39962, 2026.

\bibitem[Liang et~al.(2025)Liang, Li, Zhou, Song, Yu, Du, Mi, and
  Yu]{liang2025clue}
Zhenwen Liang, Ruosen Li, Yujun Zhou, Linfeng Song, Dian Yu, Xinya Du, Haitao
  Mi, and Dong Yu.
\newblock {CLUE}: Non-parametric verification from experience via hidden-state
  clustering.
\newblock \emph{arXiv preprint arXiv:2510.01591}, 2025.

\bibitem[Lightman et~al.(2024)Lightman, Kosaraju, Burda, Edwards, Baker, Lee,
  Leike, Schulman, Sutskever, and Cobbe]{lightman2024let}
Hunter Lightman, Vineet Kosaraju, Yuri Burda, Harrison Edwards, Bowen Baker,
  Teddy Lee, Jan Leike, John Schulman, Ilya Sutskever, and Karl Cobbe.
\newblock Let's verify step by step.
\newblock In \emph{International Conference on Learning Representations}, pages
  39578--39601, 2024.

\bibitem[Madry et~al.(2018)Madry, Makelov, Schmidt, Tsipras, and
  Vladu]{madry2017towards}
Aleksander Madry, Aleksandar Makelov, Ludwig Schmidt, Dimitris Tsipras, and
  Adrian Vladu.
\newblock Towards deep learning models resistant to adversarial attacks.
\newblock In \emph{International Conference on Learning Representations}, 2018.

\bibitem[Miao et~al.(2020)Miao, Liang, and Su]{miao2020diverse}
Shen-Yun Miao, Chao-Chun Liang, and Keh-Yih Su.
\newblock A diverse corpus for evaluating and developing {English} math word
  problem solvers.
\newblock In \emph{Proceedings of the 58th Annual Meeting of the Association
  for Computational Linguistics}, pages 975--984, 2020.

\bibitem[Pan et~al.(2026)Pan, Chen, and Steinhardt]{pan2026latentqa}
Alexander Pan, Lijie Chen, and Jacob Steinhardt.
\newblock {LatentQA}: Teaching {LLMs} to decode activations into natural
  language.
\newblock In \emph{International Conference on Learning Representations}, pages
  78726--78759, 2026.

\bibitem[Patel et~al.(2021)Patel, Bhattamishra, and Goyal]{patel2021nlp}
Arkil Patel, Satwik Bhattamishra, and Navin Goyal.
\newblock Are {NLP} models really able to solve simple math word problems?
\newblock In \emph{Proceedings of the 2021 Conference of the North American
  Chapter of the Association for Computational Linguistics: Human Language
  Technologies}, pages 2080--2094, 2021.

\bibitem[Qiu et~al.(2026)Qiu, Yu, Jiang, Liu, Huang, Chu, Yu, McAuley, and
  Wu]{qiu2026does}
Jingjie Qiu, Sheldon Yu, Xunyi Jiang, Jiajin Liu, Zihan Huang, Zhendong Chu,
  Tong Yu, Julian McAuley, and Junda Wu.
\newblock Does latent chain-of-thought actually reason? a single, unifying
  taxonomy for latent reasoning.
\newblock \emph{arXiv preprint arXiv:2610.21182}, 2026.

\bibitem[Ramjee(2026)]{ramjee2026ulterior}
Sharan Ramjee.
\newblock Ulterior motives: Detecting misaligned reasoning in continuous
  thought models.
\newblock \emph{arXiv preprint arXiv:2604.23460}, 2026.

\bibitem[Romano et~al.(2020)Romano, Sesia, and Cand{\`e}s]{romano2020adaptive}
Yaniv Romano, Matteo Sesia, and Emmanuel~J. Cand{\`e}s.
\newblock Classification with valid and adaptive coverage.
\newblock In \emph{Advances in Neural Information Processing Systems}, pages
  3581--3591, 2020.

\bibitem[Shen et~al.(2025)Shen, Yan, Zhang, Hu, Du, and He]{shen2025codi}
Zhenyi Shen, Hanqi Yan, Linhai Zhang, Zhanghao Hu, Yali Du, and Yulan He.
\newblock {CODI}: Compressing chain-of-thought into continuous space via
  self-distillation.
\newblock In \emph{Proceedings of the 2025 Conference on Empirical Methods in
  Natural Language Processing}, pages 677--693, 2025.

\bibitem[Snell et~al.(2025)Snell, Lee, Xu, and Kumar]{snell2024scaling}
Charlie Snell, Jaehoon Lee, Kelvin Xu, and Aviral Kumar.
\newblock Scaling {LLM} test-time compute optimally can be more effective than
  scaling parameters for reasoning.
\newblock In \emph{International Conference on Learning Representations}, pages
  10131--10165, 2025.

\bibitem[Tibshirani et~al.(2019)Tibshirani, Foygel~Barber, Cand{\`e}s, and
  Ramdas]{tibshirani2019conformal}
Ryan~J. Tibshirani, Rina Foygel~Barber, Emmanuel~J. Cand{\`e}s, and Aaditya
  Ramdas.
\newblock Conformal prediction under covariate shift.
\newblock In \emph{Advances in Neural Information Processing Systems}, pages
  2526--2536, 2019.

\bibitem[Torrielli et~al.(2026)Torrielli, Schneider-Kamp, and
  Poech]{torrielli2026confidence}
Federico Torrielli, Peter Schneider-Kamp, and Lukas~Galke Poech.
\newblock Confidence and calibration of activation oracles for reliable
  interpretation of language model internals.
\newblock \emph{arXiv preprint arXiv:2605.26045}, 2026.

\bibitem[Turpin et~al.(2023)Turpin, Michael, Perez, and
  Bowman]{turpin2023language}
Miles Turpin, Julian Michael, Ethan Perez, and Samuel Bowman.
\newblock Language models don't always say what they think: Unfaithful
  explanations in chain-of-thought prompting.
\newblock In \emph{Advances in Neural Information Processing Systems}, pages
  74952--74965, 2023.

\bibitem[Vovk et~al.(2005)Vovk, Gammerman, and Shafer]{vovk2005algorithmic}
Vladimir Vovk, Alexander Gammerman, and Glenn Shafer.
\newblock \emph{Algorithmic learning in a random world}.
\newblock Springer New York, 2005.

\bibitem[Wang et~al.(2026)Wang, Huang, Sun, Wei, and Wu]{wang2026out}
Chenxi Wang, Ruiyang Huang, Jiayan Sun, Lei Wei, and Yifan Wu.
\newblock Out of sight, not out of mind: Unveiling latent attack in
  latent-based multi-agent systems.
\newblock In \emph{Proceedings of the 2026 Conference on Empirical Methods in
  Natural Language Processing}, pages 47668--47696, 2026.

\bibitem[Wang et~al.(2024)Wang, Li, Shao, Xu, Dai, Li, Chen, Wu, and
  Sui]{wang2024math}
Peiyi Wang, Lei Li, Zhihong Shao, Runxin Xu, Damai Dai, Yifei Li, Deli Chen, Yu
  Wu, and Zhifang Sui.
\newblock {Math-Shepherd}: Verify and reinforce {LLMs} step-by-step without
  human annotations.
\newblock In \emph{Proceedings of the 62nd Annual Meeting of the Association
  for Computational Linguistics (Volume 1: Long Papers)}, pages 9426--9439,
  2024.

\bibitem[Wei et~al.(2022)Wei, Wang, Schuurmans, Bosma, Ichter, Xia, Chi, Le,
  and Zhou]{wei2022chain}
Jason Wei, Xuezhi Wang, Dale Schuurmans, Maarten Bosma, Brian Ichter, Fei Xia,
  Ed~H. Chi, Quoc~V. Le, and Denny Zhou.
\newblock Chain-of-thought prompting elicits reasoning in large language
  models.
\newblock In \emph{Advances in Neural Information Processing Systems}, pages
  24824--24837, 2022.

\bibitem[Yang et~al.(2025)Yang, Li, Yang, Zhang, Hui, Zheng, Yu, Gao, Huang,
  Lv, et~al.]{yang2025qwen3}
An Yang, Anfeng Li, Baosong Yang, Beichen Zhang, Binyuan Hui, Bo Zheng, Bowen
  Yu, Chang Gao, Chengen Huang, Chenxu Lv, et~al.
\newblock {Qwen3} technical report.
\newblock \emph{arXiv preprint arXiv:2505.09388}, 2025.

\bibitem[Yang et~al.(2026)Yang, Yu, Chen, Shen, and Gan]{yang2026machine}
Zeyuan Yang, Xueyang Yu, Delin Chen, Maohao Shen, and Chuang Gan.
\newblock Machine mental imagery: Empower multimodal reasoning with latent
  visual tokens.
\newblock In \emph{Proceedings of the IEEE/CVF Conference on Computer Vision
  and Pattern Recognition}, pages 33510--33520, 2026.

\bibitem[Zhang and Emu(2026)]{zhang2026latent}
Huixiang Zhang and Mahzabeen Emu.
\newblock Do latent channels actually communicate? a causal audit of latent
  multi-agent {LLM}.
\newblock \emph{arXiv preprint arXiv:2607.26773}, 2026.

\bibitem[Zhao et~al.(2026{\natexlab{a}})Zhao, He, Wang, Payani, Li, and
  Du]{zhao2026universal}
Haiyan Zhao, Zirui He, Guanchu Wang, Ali Payani, Yingcong Li, and Mengnan Du.
\newblock Universal activation verbalizer: A unified framework for cross-model
  activation explanation.
\newblock In \emph{Proceedings of the 2026 Conference on Empirical Methods in
  Natural Language Processing}, pages 11931--11959, 2026{\natexlab{a}}.

\bibitem[Zhao et~al.(2026{\natexlab{b}})Zhao, Hei, Shi, Deng, Zou, and
  Du]{zhao2026faithful}
Haiyan Zhao, Zirui Hei, Wei Shi, Huiqi Deng, Na Zou, and Mengnan Du.
\newblock Faithful activation verbalization: Reducing hallucinations in {LLM}
  representation interpretation.
\newblock \emph{arXiv preprint arXiv:2609.34033}, 2026{\natexlab{b}}.

\bibitem[Zhu et~al.(2025)Zhu, Peng, Cheng, Qu, Huang, Zhu, Wang, Xue, Zhang,
  Shan, et~al.]{zhu2025survey}
Rui-Jie Zhu, Tianhao Peng, Tianhao Cheng, Xingwei Qu, Jinfa Huang, Dawei Zhu,
  Hao Wang, Kaiwen Xue, Xuanliang Zhang, Yong Shan, et~al.
\newblock A survey on latent reasoning.
\newblock \emph{arXiv preprint arXiv:2507.06203}, 2025.

\bibitem[Zou et~al.(2026)Zou, Qiu, Li, Yang, Tieu, Lu, Shen, Tong, Choi, He,
  Zou, Wang, and Yang]{zou2025latent}
Jiaru Zou, Ruizhong Qiu, Gaotang Li, Xiyuan Yang, Katherine Tieu, Pan Lu, Ke
  Shen, Hanghang Tong, Yejin Choi, Jingrui He, James Zou, Mengdi Wang, and Ling
  Yang.
\newblock Latent collaboration in multi-agent systems.
\newblock In \emph{Proceedings of the 43rd International Conference on Machine
  Learning}, pages 168813--168847. PMLR, 2026.

\end{thebibliography}
}
\end{document}